\documentclass[11pt,a4paper]{article}

\usepackage{graphicx} 
\usepackage{subfigure} 
\usepackage{natbib}
\bibpunct{(}{)}{;}{a}{,}{,}
\usepackage{algorithm}
\usepackage{algorithmic}
\usepackage{hyperref}

\usepackage{amssymb,amsmath,amsfonts}
\usepackage{pgf}
\usepackage{xcolor}
\usepackage{lineno}
\definecolor{lnkcol}{rgb}{0,0,0.93}
\definecolor{extcol}{rgb}{0.33,0,0.55}
\hypersetup{
    pdftoolbar=true,        % show Acrobat?s toolbar?
    pdfmenubar=true,        % show Acrobat?s menu?
    pdffitwindow=true,      % page fit to window when opened
    pdfauthor={Sylvain Chevallier, Quentin Barthelemy and Jamal Atif},     % author
    pdfsubject={Metrics for multivariate dictionaries},   
    pdfnewwindow=true,      % links in new window
    pdfkeywords={Dictionary Learning, Metric Space, Frames, Grassmannian Manifold, Packing, Compressed Sensing, Multivariate Dataset}, 
    colorlinks=true,        % false: boxed links; true: colored links
    linkcolor=lnkcol,       % color of internal links
    citecolor=lnkcol,       % color of links to bibliography
    urlcolor=extcol         % color of external links
}

\newcommand{\ra}{\ensuremath{\rightarrow \;}}
 
\newcommand{\mc}[1]{\ensuremath{\mathcal{#1}}}

\newcommand{\tmop}[1]{\ensuremath{\operatorname{#1}}}
\newcommand{\norm}[1]{\ensuremath{\parallel \!\! #1 \!\! \parallel}}
\newcommand{\bignorm}[1]{\ensuremath{\left|\!\left| #1 \right|\!\right|}}

\renewcommand{\Re}{\ensuremath{\mathbb{R}}}

\def\smallskip{\vskip\smallskipamount}
\def\medskip{\vskip\medskipamount}
\def\bigskip{\vskip\bigskipamount}
\newcommand{\Gr}{\ensuremath{\mathbf{Gr}}}
\newcommand{\St}{\ensuremath{\mathbf{St}}}
\newcommand{\Or}{\ensuremath{\mathbf{O}}}
\newcommand{\GL}{\ensuremath{\mathbf{GL}}}
\newcommand{\fone}{\ensuremath{u}}
\newcommand{\ftwo}{\ensuremath{w}}
\newcommand{\Fone}{\ensuremath{U}} % le manifold F_1
\newcommand{\Ftwo}{\ensuremath{W}} % le manifold F_2
\newcommand{\FOone}{\ensuremath{\underline{U}\,}} % base orthonormale associee a F
\newcommand{\FOtwo}{\ensuremath{\underline{W}\,}} % base orthonormale associee a F
\newcommand{\subspaceFone}{\ensuremath{\mc{U}}} % span de F_1
\newcommand{\subspaceFtwo}{\ensuremath{\mc{W}}} % span de F_2
\newcommand{\CollSubspaceFone}{\ensuremath{\mathbb{U}}}
\newcommand{\CollSubspaceFtwo}{\ensuremath{\mathbb{W}}}
\newcommand{\CollFone}{\ensuremath{\mathbf{U}}}
\newcommand{\CollFtwo}{\ensuremath{\mathbf{W}}}
\newcommand{\AnyAdef}{\ensuremath{A=\{a_\find\}_{\find\in I}}}
\newcommand{\AnyBdef}{\ensuremath{B=\{b_\find\}_{\find\in J}}}
\newcommand{\AnyA}{\ensuremath{A}}
\newcommand{\AnyB}{\ensuremath{B}}

\newcommand{\sensing}{\ensuremath{u}}
\newcommand{\Sensing}{\ensuremath{U}}
\newcommand{\ses}{\ensuremath{y}}
\newcommand{\sem}{\ensuremath{a}}
\newcommand{\DicoUni}{\ensuremath{U}}
\newcommand{\dicoUni}{\ensuremath{u}}
\newcommand{\Dico}{\ensuremath{\mathbf{U}}}
\newcommand{\dico}{\ensuremath{U}}
\newcommand{\M}{m} %M nb atoms, previously p, previously L
\newcommand{\N}{n} %N dim atom
\newcommand{\Vdim}{\ensuremath{\rho}}
\newcommand{\V}{\ensuremath{\varrho}} %V nb modalites,           previously m
\newcommand{\Q}{q} % taille du training set
\newcommand{\qind}{j} % indice du training set

\newcommand{\find}{\ensuremath{i}} % previously i
\newcommand{\paind}{\ensuremath{j}}

\newcommand{\field}{\ensuremath{\mathbb{R}}}
\newcommand{\vecspace}{\ensuremath{\mc{V}}}
\newcommand{\multivarspace}{\ensuremath{\mathbb{G}}}
\newcommand{\ripl}{\ensuremath{\delta}}
\newcommand{\dist}{\ensuremath{d}}
\newcommand{\measure}{\ensuremath{\pi}}
\newcommand{\dH}{\ensuremath{d_H}}
\newcommand{\dW}{\ensuremath{d_W}}
\newcommand{\dF}{\ensuremath{d_F}}
\newcommand{\support}[1]{\tmop{Spt}(#1)}
\newcommand{\y}{\ensuremath{y}}%u
\newcommand{\z}{\ensuremath{z}}%w
\newcommand{\Y}{\ensuremath{\underline{Y}\,}}%\FOone
\newcommand{\Z}{\ensuremath{\underline{Z}\,}}%\FOtwo
\renewcommand{\dim}{\tmop{dim}}
\newcommand{\p}{r}
\newcommand{\setmetric}[2]{\ensuremath{d_{\tmop{#1}}^{\;\tmop{#2}}}}
\newcommand{\DataUni}{\ensuremath{Y}}
\newcommand{\dataUni}{\ensuremath{y}}
\newcommand{\Data}{\ensuremath{\mathbf{Y}}}
\newcommand{\data}{\ensuremath{Y}}
\newcommand{\Coef}{\ensuremath{A}}
\newcommand{\coef}{\ensuremath{a}}
\newcommand{\datastr}{\ensuremath{\mathbf{Y}_{\tmop{straight}}}}
\newcommand{\datarot}{\ensuremath{\mathbf{Y}_{\tmop{rotation}}}}
\newcommand{\AppendixManifold}{A}

\renewenvironment{abstract}
{\centerline{\large\bf Abstract}\vspace{0.7ex}%
  \bgroup\leftskip 20pt\rightskip 20pt\small\noindent}%
{\par\egroup\vskip 0.25ex}
\newenvironment{keywords}
{\bgroup\leftskip 20pt\rightskip 20pt \small\noindent{\bf Keywords:} }%
{\par\egroup\vskip 0.25ex}
\long\def\acks#1{\vskip 0.3in\noindent{\large\bf Acknowledgments}\vskip 0.2in
\noindent #1}
\newcommand{\BlackBox}{\rule{1.5ex}{1.5ex}}  % end of proof
\newenvironment{proof}{\par\noindent{\bf Proof\ }}{\hfill\BlackBox\\[2mm]}
 
\newtheorem{theorem}{Theorem}
\newtheorem{lemma}[theorem]{Lemma} 
\newtheorem{proposition}[theorem]{Proposition}

\newtheorem{definition}[theorem]{Definition}

\title{Metrics for Multivariate Dictionaries}
\author{Sylvain Chevallier\\
  LISV - Universit\'e de Versailles Saint-Quentin\\
  \vspace{0.3cm}
  78140 Velizy, France\\
  Quentin Barth\'elemy\\
  CEA, LIST - LOAD\\
  \vspace{0.3cm}
  91191 Gif-sur-Yvette, France\\
  Jamal Atif\\
  LRI - Universit\'e Paris-Sud 11\\
  \vspace{0.3cm}
  91410 Orsay, France\\
  \texttt{sylvain.chevallier@uvsq.fr}, \texttt{quentin.barthelemy@cea.fr},\\ 
  \texttt{jamal.atif@lri.fr}\\
}

\begin{document} 

\maketitle

\begin{abstract}%
Overcomplete representations and dictionary learning algorithms kept attracting a growing interest in the machine learning community. 
This paper addresses the emerging problem of comparing multivariate overcomplete representations. 
Despite a recurrent need to rely on a distance for  learning or assessing multivariate overcomplete representations, no metrics in their underlying spaces have yet been proposed.
Henceforth we propose to study overcomplete representations from the perspective of frame theory and matrix manifolds.
We consider distances between multivariate dictionaries as distances between their spans which reveal to be elements of a Grassmannian manifold. 
We introduce Wasserstein-like set-metrics defined on Grassmannian spaces and study their properties both theoretically and numerically.  
Indeed a deep experimental study based on  tailored synthetic datasetsand real EEG signals for Brain-Computer Interfaces (BCI) have been conducted. 
In particular, the introduced metrics have been embedded in clustering algorithm and applied to BCI Competition IV-2a for dataset quality assessment.
Besides, a principled connection is made between three close but still disjoint research fields, namely, Grassmannian packing, dictionary learning and compressed sensing.
\end{abstract} 

\begin{keywords}
Dictionary Learning, Metrics, Frames, Grassmannian Manifolds, Packing, Compressed Sensing, Multivariate Dataset.
\end{keywords}

%%%%%%%%%%%%%%%%%%%%%%%%%%%%%%%%%%%%%%%%%%%%%%%%%%%%%%%%%%%%%%%%%%%

\section{Introduction}
\label{sec:intro}

A classical question in machine learning is how to choose a good feature space to analyze the input data.
One possible and elegant response is to learn this feature space over mild hypotheses.
This approach is known as dictionary learning and follows the dictionary-based framework which have provided several important results in the last decades \citep{MEY95,MAL99,TRO04,CAN06a,GRI10,JEN12}.
In dictionary-based approaches, an expert selects a specific family of basis functions, called atoms, known to capture important features of the input data, for example wavelets \citep{MAL99} or curvelets \citep{CAN04}.
When no specific expert knowledge is available, dictionary learning algorithms could learn those atoms from a given dataset. 
The problem is then stated as an optimization procedure usually under some sparsity constraints. 
Thanks to the pioneering works on sparsity constraint decomposition of \citet{Mallat1993a}, on the Lasso problem \citep{TIB96} and on sparse signal recovery \citep{CAN06}, efficient algorithms are available both for projection on overcomplete representations \citep{EFR04,BEC09} and dictionary learning \citep{Engan2000,Aharon2006,Mairal2010}.

Despite the very profusion of research papers dealing with dictionary-based approaches and overcomplete representations in general, except some noticeable exceptions~\citep{BAL99,SKR10}, few results have been reported on how the constructed representations should be compared, and in a more theoretical standpoint, on which topological space these representations live.

Thus, to qualitatively assess a specific dictionary learning algorithm, one has to indirectly evaluate it through a benchmark based on a task performance \citep{Engan2000,GRO07,MAI09a,SKR10}.
Meanwhile, one can find in the literature some hints for dictionaries comparison with the aim of learning assessment \citep{ALD95,Aharon2006,Vainsencher2011,SKR11}, but they fall short to define a true metric. 
A recurrent view to dictionary comparison is based on the cross-gramian matrix and is only defined on univariate dictionaries. 
In \citet{LES07}, it is called a correlogram and measures, such as precision and recall, were defined.
In \citet{ALD95}, dictionaries are compared through the analysis Gram operator induced norm.
The work of \citet{BAL99} introduces a mapping from the frame sets to a continuous functional space.
The constructed functions allow then for the definition of an equivalence class and a partial order.
Nonetheless this approach is not invariant of linear transformation between frame elements.

In harmonic analysis, overcomplete dictionaries are called frames~\citep{christensen2003introduction}. 
Frames  are tightly linked with the work of Gabor, as they were formally defined by \citet{DUF52} and later popularized by \citet{DAU86} to be widely studied as a theoretical framework for wavelets \citep{MAL99}.
They have been used in numerous domains, for example signal processing,
compression or sampling.
They have also been actively investigated for solving packing problems \citep{CON96,STR03,DHI08}.
In its original formulation by \citet{TAM30}, the packing problem in a compact metric space consists in finding the best way to embed points on the surface of a sphere such that each pair of point is separated by the largest distance as possible.
When considering the packing of subspaces, which has applications in wireless communications, it has been the topic of a very active community, see for example \citep{CAS04,KUT09,STR12}.
%zz:  use explanation from \citep{Kutyniok2009}

Of particular interest is the remarkable equivalence between the Equiangular Tight Frames design and the Grassmannian packing problem~\citep{STR03}. This opens new exciting perspectives on exploiting the theoretical results on Grassmannian manifolds and bring them to the context of frame theory and inherently overcomplete dictionaries.

Several metrics have been proposed in Grassmannian spaces, such as the chordal distance \citep{CON96,HAM08}, the projection-norm \citep{EDE99} or the Binet-Cauchy distance \citep{WOL03,VIS04}.
These distances have been used to define a reproducing kernel Hilbert space, for SVM-based approaches or to find the best packing with a given subspace, and were also used in image processing, see \citet{LUI12} for a review.

The Grassmannian spaces define a natural framework to work with multicomponent dataset.
This opens new approaches to account for finding spatial relations in multivariate time-series or in multichannel images. 
Previous works on dictionary learning in a multivariate context are applied on multimodal audio-visual data \citep{MON07, Monaci2009}, electro-cardiogram signals \citep{Mailhe2009a}, color images \citep{Mairal2008}, hyperspectral images \citep{MOU09}, stereo images \citep{Tosic2011a}, 2D spatial trajectories \citep{Barthelemy2012} and electro-encephalogram (EEG) signals \citep{Barthelemy2013a}.

\textbf{Main contribution} of the paper is to introduce metrics suited to multivariate dictionaries, which are invariant to linear combinations (see Section~\ref{sec:metrics}).
To define these metrics, intermediate results were necessary. In particular, transport metrics are defined over Grassmannian manifolds. A principled connection has been made between multivariate dictionaries, frame theory and Grassmannian manifolds by exploiting the connection between the frame design problem and the Grassmannian packing (Section~\ref{sec:sparseframe}).

In the sequel, we first begin by recalling some definitions about Grassmannian manifolds and their metrics in Section~\ref{sec:prelim}, supplementary material on matrix manifolds are given in Appendix~\AppendixManifold. In Section~\ref{sec:frames}, necessary material from frame theory is recalled. An original formalized connection between Grassmannian frames, compressed sensing and dictionary learning is detailed in Section~\ref{sec:sparseframe}. In Section~\ref{sec:metrics}, distances between overcomplete representations are constructed as Wasserstein-like set metrics where ground metrics are those defined on Grassmannian manifolds. 
In Section~\ref{sec:multivardla}, dictionary learning principle is detailed for multivariate data, with different models.
Assessment on synthetic data is presented in Section~\ref{sec:simu_data} with experimental results for several multivariate dictionary learning algorithms, that is overcomplete representations on multichannel dataset. 
Proposed metrics allow to estimate the empirical convergence of dictionary learning algorithms and thus to evaluate the contribution of these different algorithms.
Experiences on real dataset are shown in Section~\ref{sec:reseeg}, exploiting the overcomplete representations to assess the quality of training dataset through unsupervised analyses.
A well known multivariate dataset is investigated, namely the Brain-Computer Interface Competition IV-2a.
An evaluation of the subject variability and of the difference between the training and testing datasets is produced, hence offering novel insights on the dataset consistency.
Section~\ref{sec:ccl} concludes this paper and points out some future research directions.

%%%%%%%%%%%%%%%%%%%%%%%%%%%%%%%%%%%%%%%%%%%%%%%%%%%%%%%%%%%%%%%%%%%

\section{Grassmannian Manifolds and their Metrics}
\label{sec:prelim}

This section recalls some definitions from algebraic geometry that will be helpful to characterize the metrics associated with Grassmannian manifolds.
Some basic definitions, including the Grassmannian manifold one, is given in Appendix~\AppendixManifold.

\subsection{Preliminaries}

We consider here an $\N$-dimensional vector space $\vecspace$ with inner product $\langle \cdot, \cdot \rangle$ defined over the field $\field$. % and, without loss of generality, we will assume here that $\field=\Re$.
We will denote by $u$, $w$ the elements (vectors) of $\vecspace$ and by $\Fone$, $\Ftwo$ the matrices of dimension $\N\times\Vdim$ defined as $\Fone=[\fone_1, \ldots, \fone_{\Vdim}]$, $\Ftwo=[\ftwo_1, \ldots, \ftwo_{\Vdim}]$.
The transpose of a vector $\fone$ (or a matrix $\Fone$) is denoted $\fone^T$ ($\Fone^T$). 
% Unless stated otherwise, the vector space $\vecspace$ is endowed with the usual inner product $\langle \mathbf{x}, \mathbf{y} \rangle = \mathbf{y}^T\mathbf{x}$.
The inner product induces the $\ell_2$ norm $\norm{\fone}^2_2=\langle \fone, \fone \rangle$.
The Frobenius norm is defined as $\norm{\Fone}^2_F=\tmop{trace}(\Fone^T\Fone)$ and its associated inner product is defined as $\langle \Fone,\Ftwo \rangle_F=\tmop{trace}(\Ftwo^T \Fone)$.
The subspace spanned by $\Fone$ is denoted as $\subspaceFone$ and its dimension (rank of $\Fone$) as $\V$, thus $\tmop{dim}(\subspaceFone)=\tmop{rank}(\Fone)=\V$.

\subsection{Principal Angles}
\label{sec:pa}

\begin{figure}[hbt]
  \centering
  \resizebox{0.45\linewidth}{!}{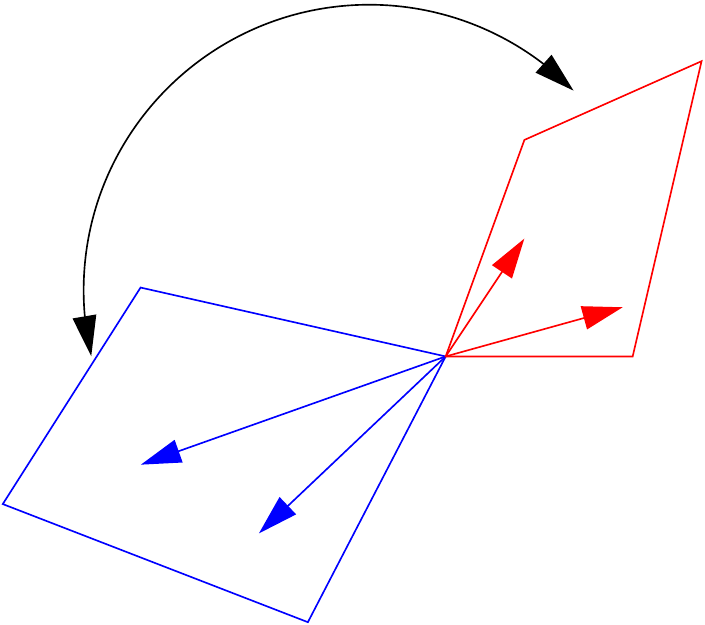}  
  \caption{Illustration of the principal angles between two subspaces $\subspaceFone$ and $\subspaceFtwo$. In this example, $\Vdim=\V=2$, $\theta_1$ is null and $\theta_2$ is plotted in the figure.}
  \label{fig:principalangles}
\end{figure}

A central notion to characterize the distance between two subspaces is the \emph{principal angles}, which is easy to interpret as shown on Figure~\ref{fig:principalangles}.

\begin{definition}
The principal angles $0 \leqslant \theta_1 \leqslant \ldots \leqslant \theta_\V \leqslant \frac{\pi}{2}$ between two subspaces $\subspaceFone$ and $\subspaceFtwo$ are defined recursively by
\begin{equation*}
  \label{eq:pa}
  \begin{split}
    \cos \theta_k &= \max_{\y_k \in \subspaceFone} \; \max_{\z_k \in \subspaceFtwo} \ \y_k^{T}\z_k\\
    & \text{s.t. } \norm{\y_k}_2 = 1, \; \norm{\z_k}_2 = 1, \\
    & \text{and } \ \y_k^{T}\y_\paind = 0,  \; \z_k^{T}\z_\paind = 0, \; \paind=0, \ldots, k-1 \ .
  \end{split}
\end{equation*}
\end{definition}
It should be noticed that principal angles are also defined when the dimension of the two subspaces are different, for example $\p= \dim \subspaceFtwo \geqslant  \dim \subspaceFone = \V \geqslant 1$, in which case the maximum number of principal angles is equal to the dimension of the smallest subspace.
%: $\V = \min (\dim \mc{A}, \dim \mc{B})$.

A more computationally tractable definition of the principal angles relies on the singular value decomposition of two bases $\AnyAdef$ spanning $\subspaceFone$ and $\AnyBdef$ spanning $\subspaceFtwo$, with $I$ and $J$ two indexing sets:
\begin{equation*}
  \label{eq:pasvd}
  \AnyA^T\AnyB = \Y \Sigma \Z^T = \Y (\cos \boldsymbol\theta) \Z^T \ ,
\end{equation*}
where $\Y=[\y_1, \ldots, \y_\V]$, $\Z=[\z_1, \ldots, \z_\V]$.
We denote by $\boldsymbol\theta$ the $\V$-vector formed by the principal angles $\theta_k$, $k = 1, \ldots, \V$.
Here $\cos\boldsymbol\theta$ is the diagonal matrix formed by $\cos \theta_1,\;\ldots,\;\cos \theta_\V$ along the diagonal. %$ = \tmop{diag}(\cos \theta_1, \ldots, \cos \theta_\V)$.
The $\cos\boldsymbol\theta$ is also known as the principal correlations or the canonical correlations \citep{HOT36,GOL95}.

\subsection{Metrics on Grassmannian Manifolds}
\label{sec:du}

In the following, we will review some of the most known Grassmannian metrics. 
Even if similar reviews could be found in~\citet{EDE99} or in~\citet{DHI08}, the one presented here is more detailed and more complete.
Most of the Grassmannian metrics are relying on principal angles.

A metric, defined for an arbitrary non empty set $X$, is a function $d : X \times X \ra \Re^+ = [0, \infty )$ which verify these three axioms:
\begin{description}
\item[(A1) Separability] $d(x,y)=0$ \emph{iff} $x=y$, $\forall x,y \in X$ , 
\item[(A2) Symmetry] $d(x,y)=d(y,x)$,  $\forall x,y \in X$ ,
\item[(A3) Triangle inequality] $d(x,y) \leqslant d(x,z)+d(z,y)$, for all $x,y,z \in X$ .
\end{description}
% When relaxing the identity of indiscernibles axiom, i.e. when some points of the associated metric space are not distinguishable, the considered distance is a pseudometric.
% A pseudometric verifies the symmetry and triangle inequality as well as:
% \begin{description}
% \item[Identity] $d(x,x)=0$, $\forall x \in X$
% \end{description}
% It could generate a pseudometric space, also call gauge space which is a uniform space. 
% As a metric space, a gauge space could be proven to be continuous or complete.

% Some useful properties of a metric for this study are defined hereafter:
% \begin{description}
% \item[Translation invariance] $d(x+z,y+z)=d(x,y), \forall x,y,z \in X$% with $X$ an abelian group.
% \item[Homogeneity] $d(\alpha x, \alpha y) = |\alpha|d(x,y)$, $\forall x,y\in X$ and $\alpha \in \field$%, with $X$ a commutative ring.
% \item[Scaling invariance] $d(x,y*\alpha)=d(x,y)$, $\forall x,y\in X$ and $\alpha \in \field$%, with $X$ a commutative ring.
% \end{description}
% Not interesting here.
% \begin{itemize}
% \item Lipschitz : $d_\rho(x, A) = \inf\{\rho(a,x) | a\in A\}, x\in X$\\
% $d_\rho(x,A)\leqslant d_\rho(y,A)+\rho(x,y), \forall x,y\in X$
% \item Intrinsic metric: for metric space $(E,d)$, $\gamma : [0,1] \ra E$, with $\gamma(0)=x$ and $\gamma(1)=y$, $x,y\in E$. If $d_I(x,y)=\inf\{L(\gamma)\}=d(x,y), \forall x,y\in E$, it is a path metric space, with the length of the path $\gamma$ is $\L(\gamma)=\sup_{0=t_0<t_1<\ldots<t_n=1}\sum_{i=0}^{n-1}d(f(t_i),f(t_{i+1}))$ and $n$ is unbounded.
% \end{itemize}

Let $\Gr(\V,\N)$ be a Grassmannian manifold, as defined in Appendix~\AppendixManifold .
Let $\subspaceFone,\subspaceFtwo$ be two elements of $\Gr(\V,\N)$ and let $\{\theta_1, \ldots, \theta_\V\}$ be their associated principal angles.

The metrics on Grassmannian manifolds rely on principal angles and a general remark applying to all metrics based on principal angles is that they are significant if and only if $\subspaceFone$ and $\subspaceFtwo$ are defined in the same space $\vecspace$. %$\Re^\N$.
Furthermore, if $\subspaceFone$ and $\subspaceFtwo$ share basis vectors, an important proportion of the principal angles could be null.
Assuming that $\Fone$ and $\Ftwo$, such that $\tmop{span}(\Fone)=\subspaceFone$ and $\tmop{span}(\Ftwo)=\subspaceFtwo$, forms separated bases the maximum number of non-null angles is $\min(\V, \N-\V)$.

\paragraph{1. Geodesic distance.} A first metric to consider is the geodesic distance or arc length first introduced by~\citet{WON67}.
\begin{definition}
Let $\subspaceFone$ and $\subspaceFtwo$ be two subspaces and ${\boldsymbol\theta}=[\theta_1, \cdots, \theta_\V]$ be the $\V$ principal angles between $\subspaceFone$ and $\subspaceFtwo$, where $\V$ is the dimension of the smallest subspace. 
The geodesic distance between $\subspaceFone$ and $\subspaceFtwo$ is:
\begin{equation}
  \label{eq:geodwong}
  d_{\tmop{g}} (\subspaceFone,\subspaceFtwo) = \,\norm{\boldsymbol\theta}_2 \,= \left( \sum_{k=1}^\V \theta^2_k \right)^{\frac{1}{2}} \ .
\end{equation}
\end{definition}
Nonetheless this distance is not differentiable everywhere \citep{CON96}.
It takes values between zero and $\frac{\pi\sqrt{\V}}{2}$.

\paragraph{2. Chordal distance.} Probably the most known Grassmannian metric~\citep{GOL96,CON96,WAN06}. 
% As the chordal distance is a metric on the Grassmannian, it is invariant to all transformations induced by $\GL(\V)$ and one could rely on orthogonal bases $\FOone$ associated with the considered subspace $\Fone$
%It is obtained by embedding the $\Gr(\V,\N)$ in the set of $\N \times \N$ projection matrices of rank $\V$.% \citep{EDE99}:

\begin{definition}
Let $\subspaceFone$ and $\subspaceFtwo$ be two subspaces and ${\boldsymbol\theta}=[\theta_1, \cdots, \theta_\V]$ be the $\V$ principal angles vector between $\subspaceFone$ and $\subspaceFtwo$, where $\V$ is the dimension of the smallest subspace. 
The chordal distance is defined as:
\begin{equation}
  \label{eq:chordaldef}
  d_{\tmop{c}}(\subspaceFone,\subspaceFtwo) = \,\norm{\sin \boldsymbol\theta}_2\, = \left(\sum_{k=1}^\V \sin^2 \theta_k\right)^{\frac{1}{2}} \ .
\end{equation}
\end{definition}

\begin{proposition}\textnormal{\textbf{\citep{DHI08}}}
The chordal distance can be reformulated in a more computationally effective way as:
\begin{equation*}
  \label{eq:newchordal}
  d_{\tmop{c}}(\subspaceFone,\subspaceFtwo) =  \left( \V - \norm{\FOone^T\FOtwo}^2_F\right)^{\frac{1}{2}} \ ,
\end{equation*}
where the columns of $\FOone$ and $\FOtwo$ are the orthonormal bases for $\subspaceFone$ and $\subspaceFtwo$, that is $\FOone^T\FOone = I_\V$ and $\tmop{span}\,(\FOone) = \subspaceFone$. %the subspaces of
\end{proposition}

The chordal distance approximates the geodesic distance when the planes are close and its square is differentiable everywhere. 

The chordal distance could be seen as the embedding of an element $\subspaceFone$ of $\Gr(\V,\N)$ with its projection matrix $\FOone \FOone^T$.% where $\FOone$ is an orthonormal basis for the subspace $\subspaceFone$.
The embedding could be formalized as:
\begin{align*}
	\Pi_c :& \;\Gr(\V,\N) \ra \Re^{\N\times\N}
	\\
	& \; \subspaceFone \mapsto \FOone\FOone^T \ .
	\label{eq:profChordal}
\end{align*}
Thus the Grassmannian element is embedded in the set of orthogonal projection matrices of rank $\V$.
The chordal distance could thus be expressed as:
\begin{equation*}
d_c(\subspaceFone,\subspaceFtwo)=\frac{1}{\sqrt{2}}\norm{\FOone\FOone^T - \FOtwo\FOtwo^T}_F \ .
\end{equation*}

% In \citep{CON96}, the authors call $\FOone$ and $\FOtwo$ generator matrices, then they defined the projection matrices $\FOone^T\FOone$ and $\FOtwo^T\FOtwo$, two $\V-$by$-\V$ symmetric idempotent matrix. Thus 
% \begin{align*}
% d_{c}(\Fone,\Ftwo) &= \left(\V - \tmop{trace}(\FOone^T\FOone \FOtwo^T\FOtwo)\right)^{1/2}\\
% &= \frac{1}{\sqrt{2}}\norm{\FOone^T\FOone-\FOtwo^T\FOtwo}_F  
% \end{align*}

Instead of using the Frobenius norm on the embedded Grassmannian, it is possible to use the 2-norm.
The chordal 2-norm distance writes as:
\begin{equation*}
  d_{\tmop{c2}}(\subspaceFone,\subspaceFtwo) =\, \norm{\FOone\FOone^T - \FOtwo\FOtwo^T}_2 = \,\norm{\sin \boldsymbol\theta}_\infty\, = \left(\sin^2 \theta_\V\right)^{\frac{1}{2}} \ ,
\end{equation*}
which fails to be a metric because of (A1), it is thus a pseudo-metric.

A variation of the chordal metric is the projection metric, which starts with different assumptions but yield very similar results.
The projection metric is the result of embedding $\Gr(\V,\N)$ in the vector space $\Re^{\N \times \V}$:
\begin{equation*}
  \label{eq:projF}
  \begin{split}
    d_{\tmop{proj}}(\subspaceFone,\subspaceFtwo) &= \,\norm{\FOone \Y-\FOtwo \Z}_F\\
    &=\bignorm{2 \sin \left(\frac{\boldsymbol\theta}{2}\right)}_2 = 2 \left( \sum_{k=1}^\V\sin^2\frac{\theta_k}{2}\right)^{\frac{1}{2}} \ .
  \end{split}
\end{equation*}
The projection distance is very similar to the chordal one: it is the minimum distance between all the possible representations of the two subspaces $\subspaceFone$ and $\subspaceFtwo$ \citep{CHI03}, which could be reformulated  in term of the following minimization problem:
\begin{align*}
  \label{eq:minproj2F}
  d_{\tmop{proj}}(\subspaceFone,\subspaceFtwo) = & \min_{\underline{R}_1, \underline{R}_2 \in \Or(\V)} \norm{\FOone \underline{R}_1- \FOtwo \underline{R}_2}_F\\
  = & \norm{\FOone\Y-\FOtwo\Z}_F \ .
\end{align*}
where $\Or(\V)$ is the orthogonal group.

As for the chordal distance, it is possible to rely on the 2-norm instead of the Frobenius norm for the projection metric, giving also the following pseudo-metric:
\begin{equation*}
  \label{eq:proj2}
  \begin{split} 
  d_{\tmop{proj}2}(\subspaceFone,\subspaceFtwo) &= \,\norm{\FOone \Y- \FOtwo \Z}_2\\
  &= \bignorm{2 \sin \frac{\boldsymbol\theta}{2}}_\infty = 2 \sin^2 \left(\frac{\theta_\V}{2}\right)  \ .
  \end{split}
\end{equation*}

%zz: Add interpretation of the chordal from \citep{ABS04}.

\paragraph{3. Fubini-Study distance.} The Fubini-Study distance is derived via the Pl\"ucker embedding of $\Gr(\V,\N)$ into the projective space $\mathbf{P}\left(\wedge^\V\left(\Re^\N\right)\right)$ (taking the wedge product between all elements of $\subspaceFtwo$), then taking the Fubini-Study metric \citep{KOB69}. 

\begin{definition}
Let $\subspaceFone$ and $\subspaceFtwo$ be two subspaces and ${\boldsymbol\theta}=[\theta_1, \cdots, \theta_\V]$ be the $\V$ principal angles vector between $\subspaceFone$ and $\subspaceFtwo$, where $\V$ is the dimension of the smallest subspace.
The Fubini-Study distance is defined as: 
\begin{equation}
  \label{eq:fubinistudy}
  d_{\tmop{fs}} (\subspaceFone,\subspaceFtwo) =  \arccos \left( \prod_{k=1}^\V \theta_k \right) \ .
\end{equation}
\end{definition}

The Fubini-Study distance can be computed efficiently as:
\begin{equation*}
d_{\tmop{fs}} (\subspaceFone,\subspaceFtwo)= \arccos(\det(\FOone^T\FOtwo)) \ . %\nonumber
\end{equation*}

In \cite{CON96}, the authors also argued that the Pl\"ucker embedding does not give a way to realize a geodesic or a chordal distance as an Euclidean distance.
%zz: \textcolor{red}{ajouter ref Mumford.}

\paragraph{4. Spectral distance.} Introduced in~\citet{DHI08}, the spectral distance is used to promote specific configurations of subspaces in packing problems.

\begin{definition}
Let $\subspaceFone$ and $\subspaceFtwo$ be two subspaces and ${\boldsymbol\theta}=[\theta_1, \cdots, \theta_\V]$ be the $\V$ principal angles vector between $\subspaceFone$ and $\subspaceFtwo$, where $\V$ is the dimension of the smallest subspace. 
The spectral distance is defined as: 
\begin{equation}
  \label{eq:dspec}
  \begin{split}
    d_{\tmop{s}}(\subspaceFone,\subspaceFtwo) & = \min_k \sin\theta_k \\
    & = \left( 1 - \norm{\FOone^T\FOtwo}^2_{2,2} \right)^{\frac{1}{2}}  \ ,
  \end{split}
\end{equation}
where the spectral norm $\norm{X}_{2,2}$ returns the largest singular value of $X$.
\end{definition}

This distance allows to solve the packing problem with equi-isoclinic configurations, that is when all principal angles formed by all pairs of subspaces are equal.
Nonetheless this distance, defined in Equation~\eqref{eq:dspec}, is not a metric. % as the separability axiom (A1) and the triangle inequality axiom (A3) are not verified.

\paragraph{5. Binet-Cauchy distance.} A last metric on $\Gr(\V,\N)$ is the Binet-Cauchy distance which is the product of principal angles or canonical correlations \citep{WOL03,VIS04}. 

\begin{definition}
Let $\subspaceFone$ and $\subspaceFtwo$ be two subspaces and ${\boldsymbol\theta}=[\theta_1, \cdots, \theta_\V]$ be the $\V$ principal angles vector between $\subspaceFone$ and $\subspaceFtwo$, where $\V$ is the dimension of the smallest subspace. 
The Binet-Cauchy distance is defined as: 
\begin{equation}
  \label{eq:binetcauchy}
  d_{\tmop{bc}}(\subspaceFone,\subspaceFtwo) = \big(1 - \prod_{k=1}^\V \cos^2 \theta_k \big)^{\frac{1}{2}} \ .
\end{equation}
\end{definition}

This distance was introduced to enhance the computational efficiency and the numerical stability of the Canonical Correlation Analysis (CCA) based kernel approaches.
%This metric is tightly linked with the Fubini-Study as shown in \citep[Eq.~4.7]{BER97}.

All these distances are summarized in the Table~\ref{tab:grassdist} with their definition.

\begin{table}
  \centering
  \begin{tabular}{|l|c|c|c|}
  	\hline Distance & Definition \rule[-7pt]{0pt}{20pt} & Metric & DE \\ \hline \hline
    Geodesic & $d_{\tmop{g}}(\subspaceFone,\subspaceFtwo) = \norm{\boldsymbol\theta}_2$ \rule[-7pt]{0pt}{20pt} & $\checkmark$ & $\times$ \\ \hline
    Chordal & $d_{\tmop{c}}(\subspaceFone,\subspaceFtwo) = \norm{\sin\boldsymbol\theta}_2$ \rule[-7pt]{0pt}{20pt} & $\checkmark$ & $\checkmark$ \\ \hline
    Fubini-Study & $d_{\tmop{fs}}(\subspaceFone,\subspaceFtwo) = \arccos (\det(\FOone^T\FOtwo))$ \rule[-7pt]{0pt}{20pt} & $\checkmark$ & $\checkmark$ \\ \hline
    Spectral &  $d_{\tmop{s}}(\subspaceFone,\subspaceFtwo) = \left( 1 - \norm{\FOone^T\FOtwo}^2_{2,2} \right)^{\frac{1}{2}}$ \rule[-7pt]{0pt}{20pt} & $\times$ & $\times$ \\ \hline
    Binet-Cauchy & $d_{\tmop{bc}}(\subspaceFone,\subspaceFtwo) = \big(1 - \prod_k \cos^2 \theta_k \big)^{\frac{1}{2}}$ \rule[-7pt]{0pt}{20pt} & $\checkmark$ & $\checkmark$\\ \hline
  \end{tabular}
  \caption{Distances on Grassmannian manifolds $\subspaceFone$ and $\subspaceFtwo$, DE stands for differentiable everywhere.}
  \label{tab:grassdist}
\end{table}

%%%%%%%%%%%%%%%%%%%%%%%%%%%%%%%%%%%%%%%%%%%%%%%%%%%%%%%%%%%%%%%%%%%

\section{Frames}
\label{sec:frames}

We assume that the reader is familiar with frame theory and we restrict ourselves to introduce definitions and facts that are mandatory for further developments in the paper.
Please refer to \citet{christensen2003introduction} for a deeper study of the topic.

We assume that $\vecspace$, together with its inner product and the induced norm, is a separable Hilbert space.
The indexing set $I$ is in $\mathbb{N}$ as we will consider, without loss of generality, only finite-dimensional frames.

\begin{definition}
  A family of elements $\Fone=\{\fone_\find\}_{\find\in I}$ is a frame for $\vecspace$ if there exist $0 < b_1 \leqslant b_2 \leqslant \infty$, such that for all $\ftwo \in \vecspace$:
  \begin{equation}
    \label{eq:framedef}
    b_1 \norm{\ftwo}^2 \leqslant \sum_{\find \in I} |\langle \ftwo,\fone_\find\rangle|^2 \leqslant b_2 \norm{\ftwo}^2 \ .
  \end{equation}  
\end{definition}
If $\norm{\fone_\find}^2=1$ for all $\find$, it is a unit norm frame.
A tight frame is a frame whose bounds are equal ($b_1=b_2$) and if a frame verifies $b_1=b_2=1$, it is a Parseval frame or a normalized tight frame.
This inequality implies that the frame $\Fone$ is full rank \citep{MAL99}.

\begin{definition}
  Let $\{\fone_\find\}_{\find\in I}$ be a frame for $\vecspace$. The Bessel map, or analysis operator, is a function $T: \vecspace \ra \ell^2(I)$ defined by:
  \begin{equation*}
    \label{eq:analysis}
    T(\ftwo) = (\langle \ftwo, \fone_\find \rangle)_{\find \in I} \ .
  \end{equation*}
\end{definition}

\begin{definition}
  Let $\{\fone_\find\}_{\find\in I}$ be a frame for $\vecspace$. The pre-frame operator, or the synthesis operator, is a map $T^*: L^2(I) \ra \vecspace$ defined by:
  \begin{equation*}
    \label{eq:syn}
    T^*c = \sum_{\find  \in I} c_\find \fone_\find \ .
  \end{equation*}
\end{definition}

For an orthonormal basis, these operators are invertible and are each other inverse.
In the context of frames, it is usually not the case.
Specifically, $T^*$ is always surjective but usually not injective and $T$ is always injective but usually not surjective.

\begin{definition}
  The frame operator is a mapping $S: \vecspace \rightarrow \vecspace$ and is defined as $T^*T$, that is 
\begin{equation*}
  \label{eq:frameop}
  S(\ftwo) = \sum_{\find \in I} \langle \ftwo, \fone_\find \rangle \fone_\find \ .
\end{equation*}
\end{definition}
The frame operator is well defined, bounded, self-adjoint and positive.

\section{Sparse Coding, Dictionary Learning and Grassmannian Frames}
\label{sec:sparseframe}

Our attempt here is to establish a principled connexion between three close but still disjoint fields of research, namely Grassmannian packing, dictionary learning and compressed sensing.
In this section, without loss of generality,  we will restrict the discussion to the case of $\Vdim=1$.

\subsection{Packing Problem and Grassmannian Frames}
\label{sec:packingGrassmannian}

One way to characterize a frame is to evaluate how ``overcomplete'' it is.
The redundancy expresses this measure of overcompleteness and is defined for $\Fone=\{\fone_\find\}_{\find=1}^\M$ in a $\N$-dimensional vector space $\vecspace$ as $\frac{\M}{\N}$.
A more informative measure is the coherence (or maximum correlation), defined as the maximum of the absolute inner product between any two elements of a unit norm frame $\{\fone_\find\}_{\find=1}^\M$:
\begin{equation}
  \label{eq:maxcor}
  \mu = \max_{i,j\in\{1,\ldots,\M\},i\neq j} \left\{ \big| \langle \fone_i, \fone_j \rangle \big| \right\} \ .
\end{equation}
The constraint on unit norm frame could be removed if the inner product of Equation~\eqref{eq:maxcor} is divided by norm of the frame elements.

A fundamental result in frame theory bounds for the coherence of any frame and the bound depends only on the number of elements in the frame and the dimension of the associated space $\vecspace$.

\begin{theorem}\textnormal{\textbf{\citep{WEL74}}}
    Let $\Fone=\{\fone_\find\}_{\find=1}^\M$ be a frame for $\vecspace$, then its \emph{Welch bound} is 
\begin{equation}
  \label{eq:welch}
  \mu \geqslant \sqrt{\frac{\M-\N}{\N(\M-1)}} \ .
\end{equation}
The equality hold \emph{iff} $\Fone$ is an \emph{Equiangular Tight Frame} (ETF), $|\langle \fone_i, \fone_j \rangle|=c$ for all $i$, $j$ with $i \neq j$, for some constant $c \geqslant 0 $.
Furthermore, the equality holds only if $\M \leqslant \frac{\N(\N+1)}{2}$.
\end{theorem}
%Thus any orthonormal basis is equiangular.

In order to construct equiangular tight frames, a possible approach is to define a set of frames which minimizes the coherence for all unit norm frames with the same redundancy.
\begin{definition}\textnormal{\textbf{\citep{STR03}}}
  A sequence of vectors $\Fone=\{\fone_\find\}_{\find=1}^\M$ in $\vecspace$ is called a \emph{Grassmannian frame} if it is the solution to $\min \mu$ where the minimum is taken over all unit norm frames in $\vecspace$. %$\{f_\find\}_{\find=1}^\M$ 
\end{definition}
A unit norm frame holding the equality in Equation~\eqref{eq:welch} is called an optimal Grassmannian frame.
Note that optimal Grassmannian frames do not necessarily exist for all packing problem: a known example \citep{CON96} is the packing of 5 vectors in $\Re^3$, with $\V=1$, $\M=5$, $\N=3$, whose coherence is at best $\frac{1}{\sqrt{5}}$ while its Welch bound is $\frac{1}{\sqrt{6}}$.

The ETF and the Grassmannian frames offer a well suited tool to solve the packing problem. 
In its general formulation, for  given $\N$, $\M$ and $\V$, the goal of the packing problem is to find a set of $\V$-dimensional subspaces $\subspaceFone_\find$ in a $\N$-dimensional vector space $\vecspace$ such that (i) the minimal distance between any two of these subspaces are as large as possible, or equivalently (ii) the maximal correlation between subspaces is as small as possible.
A schematic view of the packing of a space is shown on left part of Figure~\ref{fig:packlearn}.

In the case of $\Gr(1, \N)$, investigated in \citet{STR03}, the correlation depends only on $\N$ and $\M$.
Thus subspaces are lines passing through the origin in $\vecspace$ and the angles between any two of the $\M$ lines should be as large as possible.
As maximizing the angle between lines is the same as minimizing the absolute value of the inner product of the $\fone_\find$ vectors, finding an optimal packing in $\Gr(1,\N)$ is equivalent to finding a Grassmannian frame.

\subsection{Compressed Sensing vs Grassmannian Frames}
\label{sec:csgrass}

It is interesting to see that the ETF are also linked with sparse coding and compressed sensing.
Very briefly, sparse representations and compressed sensing \citep{Donoho2006} proposes to acquire a signal $\ses$ in $\vecspace$ by collecting $\M$ linear measurements of the form $\sem_k=\langle \sensing_k, \ses \rangle + e_k$ with $1 \leqslant k \leqslant \M$ or equivalently: 
\begin{displaymath}
  \sem = \Sensing^T \ses + e \ ,
\end{displaymath}
where $\Sensing$ is a $\N \times \M$ sensing matrix, with $\M \ll \N$ and $e$ is an error term.
The compressed sensing theory states that, if $\ses$ is reasonably sparse, it is possible to recover $\ses$ by convex programming, as it is the solution to
%zz:\textcolor{red}{first put $L_0$ before explaning that it is possible to use $L_1$}
\begin{equation}
  \label{eq:CSreconst}
  \min_{\tilde{\ses} \in \vecspace} \norm{\tilde{\ses}}_1 \text{ s.t. } \norm{\sem-\Sensing^T\tilde{\ses}}_2 \leqslant \epsilon \ .
\end{equation}
In a noiseless settings, with a null vector $e$, to achieve an exact reconstruction of $\ses$, a sensing matrix should satisfy the ($K,\ripl$)-Restricted Isometry Property (RIP).
\begin{definition}
  For a $\N \times \M$ sensing matrix $\Sensing$, the $K$-restricted isometry constant $\ripl$ of $\Sensing$ is the smallest quantity such that
  \begin{displaymath}
    (1-\ripl)\norm{\ses}^2_2 \;\leqslant\; \norm{\Sensing^T \ses}^2_2 \;\leqslant (1+\ripl)\norm{\ses}^2_2
  \end{displaymath}
holds for every $K$-sparse vectors $\ses$.
\end{definition}
A matrix with a small $\ripl$ constant indicates that every subset of
$K$ or less columns is linearly independent.
Furthermore, any $\Sensing$ sensing matrix which is ($2K,\ripl$)-RIP for some $\ripl < \sqrt{2}-1$ allows to recover the signal using Equation~\eqref{eq:CSreconst} with the following error bound
\begin{equation*}
  \label{eq:CSerrbound}
  \norm{\tilde{\ses}-\ses}_2\; \leqslant C_0\frac{\norm{\tilde{\ses}-\ses}_1}{\sqrt{K}}+C_1\epsilon \ .
\end{equation*}
Matrices with small $\ripl$ are abundant and could be easily obtained.
Matrices with Gaussian or Bernoulli entries have small $\ripl$ with very high probability if the number of measurement $\M$ is of the order of $K \log (\N/K)$.
However, testing a matrix for RIP is computationally intensive, as it requires to determine singular values for $\N$-by-$K$ submatrices.

An alternative approach is to rely on ETF for designing deterministically RIP matrices. From the Theorem~2 of \citet{MIX11b}, the following lemma are defined:
\begin{lemma} \textnormal{\textbf{\citep{MIX11b}}}
  Let $\Sensing=\{\sensing_\find\}_{\find\in I}$ be a unit norm frame on $\vecspace$.
  The smallest $\ripl_{\tmop{min}}$ for which $\Sensing$ is $(K,\ripl)$-RIP is defined by
  \begin{equation*}
    \label{eq:mixon}
    \ripl_{\tmop{min}} = \max_{\mathcal{S}\subseteq \{1,\ldots,\N\},|\mathcal{S}|=K} \norm{\Sensing_{\mathcal{S}}^T \Sensing_{\mathcal{S}}-I_{\mathcal{S}}}_2 \ ,
  \end{equation*}
  where $\Sensing_{\mathcal{S}}$ is the submatrix consisting of columns of $\Sensing$ indexed by $\mathcal{S}$.
\end{lemma}

As it is computationally demanding to compute all the eigenvalues of $\mathcal{S} \times \mathcal{S}$ submatrices, the Gershgorin circle theorem is used to obtain a coarse bound linked to RIP \citep{DEV07,APP09}.
\begin{lemma}\textnormal{\textbf{\citep{MIX11b}}}
  Let $\Sensing = \{\sensing_\find\}_{\find\in I}$ be a unit norm frame, then the $K$-restricted isometry constant 
  \begin{equation*}
    \label{eq:gershgorin}
    \ripl\leqslant (K - 1) \mu \ ,
  \end{equation*}
where $\mu$ is the coherence.
\end{lemma}

Thus, $\Sensing$ is $(K,\ripl)$-RIP for $\ripl \geqslant (K-1)\mu$ and using Welch bound given in Equation~\eqref{eq:welch}.
As ETF meet the Welch bound with equality, they are $(K,\ripl)$ for $\ripl^2 \geqslant \frac{(K-1)^2(\N-\M)}{\M(\N-1)}$.
Hence, ETF are good candidates for producing deterministically RIP sensing matrices.

\subsection{Dictionary Learning vs Grassmannian Frames}
\label{sec:dlgrass}

% In dictionary learning, a dictionary is collection of atoms and is not necessarily full rank \citep{MAL99}. 
% This constraint is relaxed in the recent development of the theoretical bounds of DL.

The dictionary learning aims at capturing most of the energy of a signal database $\DataUni=\{\dataUni_\qind\}_{\qind=1}^\Q \subset \vecspace$ 
%\quentin{Est-ce que ca ne serait pas plutot: $\DataUni=\{\dataUni_\qind \subset \vecspace \}_{\qind=1}^\Q \ $ ?}\sylvain{Non, Jamal tu confirmes?}  \jamal{Non on a $\dataUni_\qind \in \vecspace$ } 
and representing it through a collection $\DicoUni = \{\dicoUni_\find\}_{\find=1}^\M$ thanks to a set of sparse coefficients $\Coef=\{\coef_\qind\}_{\qind=1}^\Q$. 
This collection, which is redundant since $\M \gg \N$, is called overcomplete dictionary.
This dictionary is not necessarily full rank, which is a difference from frames as imposed by Equation~\eqref{eq:framedef}.
Formally, the dictionary learning problem writes as:
\begin{equation*}
  \label{eq:dl}
  \begin{split}
  \min_{\DicoUni,\Coef} \norm{\DataUni-\DicoUni\Coef}_F^2 \ \text{ s.t.} \ 
  & \norm{\coef_\qind}_0\, \leqslant K,\, \qind=1,\ldots,\Q 
  \\
  \text{and} \ 
  & \norm{\dicoUni_\find}_2=1,\, \find=1,\ldots,\M  \ \ ,
  \end{split}
\end{equation*}
%where the subscript $\qind$ is added to the vector of coefficients $\coef_\qind$ to denote its link with the treated signal $\data_\qind$ and 
where $\left\| \coef_\qind \right\|_0$ is the number of nonzero elements of vector $\coef_\qind$.
This problem is tackled by dictionary learning algorithms (DLAs), in which energy representative
% characteristic, zz: try a better formulation than energy representative
patterns of the dataset are iteratively selected by a sparse approximation step, and then updated by a dictionary update step.
Different algorithms deal with this problem: see for example the method of optimal directions proposed by \citet{Engan2000} and generalized under the name iterative least-squares DLA \citep{Engan2007}, the K-SVD \citep{Aharon2006} or the online DLA \citep{Mairal2010}.

\begin{figure*}[th]
  \centering
  \pgfimage[interpolate=true,width=0.3\linewidth]{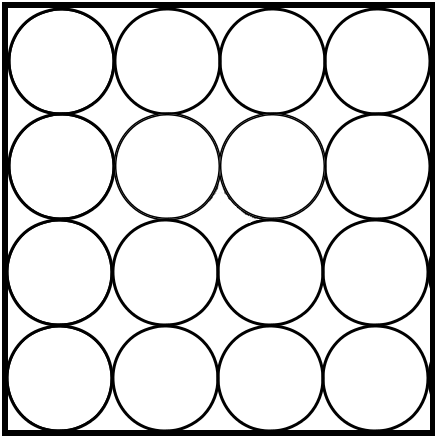}
  \pgfimage[interpolate=true,width=0.3\linewidth]{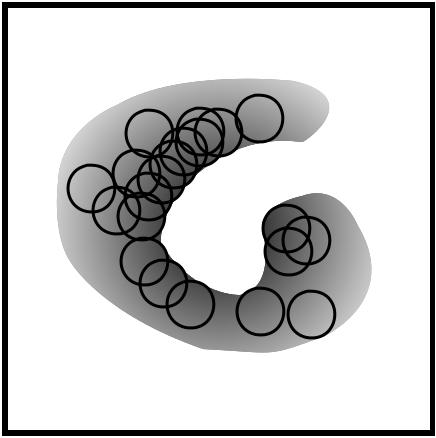}
  \pgfimage[interpolate=true,width=0.3\linewidth]{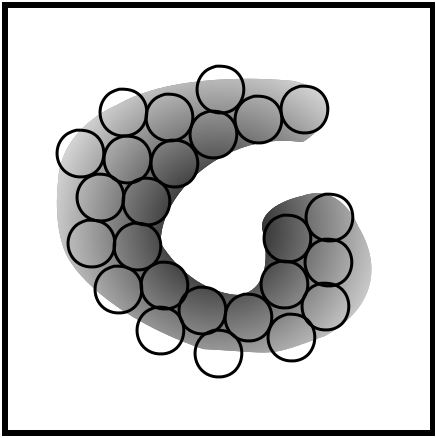}
  \caption{In the frame identification problem or packing problem (left), the frame elements (circles) are used cover the entire space under the constraint that the distance between any two elements should be maximal, being as close as possible to an ETF.
In the dictionary learning setup (center), the frame elements are chosen to minimize the reconstruction error on the energy of the dataset (represented as gray values in ``G'' shape).
An hybrid approach (right) adds coherence constraints in the dictionary learning framework.}
  \label{fig:packlearn}
\end{figure*}

Data-driven dictionary learning extracts atoms that endow most of the energy of the dataset in the underlying norm sense.
They are learned empirically to be the optimal dictionary that jointly gives sparse approximations for all of the signals of this set \citep{Tosic2011}, as illustrated on the center part of Figure~\ref{fig:packlearn}.
During learning process, no constraint is put on the dictionary elements in term of coherence.
Whereas in the packing procedure for $\M$ elements, as shown in Section~\ref{sec:packingGrassmannian}, amounts to come up with elements with maximal distance between any two of them, hence minimizing the maximal correlation.

The packing procedure is data-free, in the sense that it is independent of the data and is a pure geometrical problem, while the dictionary learning is data-driven.
An interesting view is to connect the two visions, an approach which is emerging in recent problem.
A possible approach is to constraint the packing problem on the energy of the dataset, in other words a data-constraint packing.

%zz:ETF is a span of the space vs span of the energy, seen as a constrained packing problem

An alternative point of view, as developed by \citet{Yaghoobi2009a}, is to constraint the dictionary learning with a penalty term on the coherence of the dictionary elements (see the right part of Figure~\ref{fig:packlearn}).
This could be formalized as adding the penalty term during the update of dictionary, to constraint the dictionary element to be close to ETF, as:
\begin{equation*}
  \label{eq:penyaghobi}
  \norm{\DicoUni^T\DicoUni-G}_F^2 \ \leqslant \ \epsilon  \ \  \text{with}  \ \  G=\left[
\begin{array}{ccc}
  1 & & g_{ij} \\
  & \ddots & \\
  g_{ji} & & 1 \\
\end{array}
\right] \ ,
\end{equation*}
where  $\max_{i\neq j} |g_{ij}|$ is less or equal to the Welch bound given in Equation~\eqref{eq:welch}.

As a summary, and as illustrated on Figure~\ref{fig:packlearn}, packing spans the data space equi-correlatively (left), dictionary learning spans the data energy sparsely (center), and coherence constrained dictionary learning spans the data energy sparsely and equi-correlatively (right).

%%%%%%%%%%%%%%%%%%%%%%%%%%%%%%%%%%%%%%%%%%%%%%%%%%%%%%%%%%%%%%%%%%%

\section{Metrics between multivariate dictionaries}
\label{sec:metrics}

In this section, we will use the metrics described in Section~\ref{sec:du} and summarized in Table~\ref{tab:grassdist} which act on elements of a Grassmannian as ground distance.
This ground distance has a key role for the definition of a metric between sets of points in a Grassmannian space, that is a distance between two subspaces spanned by their associated disctionaries.

\subsection{Preliminaries}
\label{sec:prelimMet}

% we have \multivarspace = \{ \subspaceFone_i\}_{i\in I}
In this section, the vector space $\vecspace$ is of dimension $\N\times\Vdim$, its elements will be denoted as $\Fone, \Ftwo \in \vecspace$. Their respective spans are $\tmop{span}(\Fone)=\subspaceFone$ and $\tmop{span}(\Ftwo)=\subspaceFtwo$. The indexed families of elements will be denoted $\CollFone=\{\Fone_i\}_{i\in I}$ and $\CollFtwo=\{\Ftwo_j\}_{j\in J}$.
Indexed families of subspaces will be denoted $\CollSubspaceFone=\{\subspaceFone_i\}_{i\in I}$ and $\CollSubspaceFtwo=\{\subspaceFtwo_j\}_{j\in J}$.

The Grassmannian manifold  $\Gr(\V,\N)$, $\V\leqslant\Vdim$, together with any of the distances $d$ discussed in Section~\ref{sec:du}, defines a metric space (or pseudo-metric space depending on the properties of the  underlying  distance). We will denote it in the sequel as $(\multivarspace,\dist)$, and when there is no confusion as $\multivarspace$. 
The following theorem characterizes the properties of Grassmannian manifolds.
\begin{theorem}\textnormal{\textbf{\citep{milnor-book}}}
\label{thm:grass}
$\Gr(\V,\N)$ is a Hausdorff, compact, connected smooth manifold of dimension $\V(\N-\V)$.
\end{theorem}
Thus, we know that $\multivarspace$ is complete and (separable) Hausdorff, and from the fact that any topological manifold is locally compact, then one can define a Borel measure  on $\multivarspace$.  We will denote by $\measure$ such a measure on the Borel $\sigma$-algebra of $\multivarspace$. The support of $\measure$ is denoted $\support{\measure}$ and is the smallest closed set on which it is concentrated, that is the minimal closed subset $X\subset \multivarspace$ such that $\measure(\multivarspace\backslash X) = 0$. The triplet $(\multivarspace, \dist,\measure)$ is  then a metric measure space. Furthermore, $\multivarspace$ is a Polish space, that is a separable completely metrizable topological space. 

% Let us denote $\multivarspace$ the Grassmannian space $\Gr(\V,\N)$ and let $\dist$ denote an elementary distance between element of $\multivarspace$.
% As $\dist$ is a metric, $(\multivarspace,\dist)$ is a metric space.
% Furthermore, $(\multivarspace,\dist)$ is complete and separable.
% Together with $\measure$, the triple $(\multivarspace, \dist,\measure)$ is metric measure space, where $\measure$ is a measure on the Borel $\sigma$-algebra of $\multivarspace$.
% The support of $\measure$ is denoted $\support{\measure}$ and is the smallest closed set on which it is concentrated.
%zz: to check
%An element of $(\multivarspace,\dist)$, i.e. a frame, is a compact subset of $\multivarspace$.
% All measures considered here are Borel and defined on a Polish space, a separable, completely metrizable topological space.
% If $\mu$ is a measure, the support of $\mu$ is denoted $\support \mu$ and is the smallest closed set on which it is concentrated.
% Thus $\multivarspace,\dist,\measure)$ is a metric measure space.

\subsection{Hausdorff Distance}
\label{sec:hausdorff}
%\sylvain{The frames are now denoted $\FrameOne$ and $\FrameTwo$}.

A classical approach to define a distance between subsets of a metric space is to rely on the Hausdorff distance. Within our context, frames are subsets of the space $\multivarspace$, and the Hausdorff distance is a first natural way to define a distance in the frame space. 

We denote by $B_r(\CollSubspaceFtwo)$ the $r$-neighborhood of a set $\CollSubspaceFtwo$ in the metric space $(\multivarspace,d)$, that is the set of points $\subspaceFone$ such that $\inf\{d(\subspaceFone,\subspaceFtwo):\subspaceFtwo\in \CollSubspaceFtwo\}<r$.

\begin{definition}
  \label{eq:hausdorff}
Let $\CollSubspaceFone=\{\subspaceFone_i\}_{i \in I}$ and $\CollSubspaceFtwo=\{\subspaceFtwo_j\}_{j\in J}$ be two subsets of the metric space $(\multivarspace,d)$. The Hausdorff distance between $\CollSubspaceFone$ and $\CollSubspaceFtwo$, denoted by $ \dH(\CollSubspaceFone,\CollSubspaceFtwo)$, is defined as
\begin{equation*}
  %\dH(\Fone,\Ftwo) := \max \left( \sup_{\fone \in \Fone} \inf_{\ftwo \in \Ftwo} \dist(\fone,\ftwo), \sup_{\ftwo \in \Ftwo} \inf_{\fone\in\Fone} \dist(\fone,\ftwo) \right)
  \dH(\CollSubspaceFone,\CollSubspaceFtwo) := \inf\{r>0:\CollSubspaceFone\subset B_r(\CollSubspaceFtwo) \text{~and~} \CollSubspaceFtwo\subset B_r(\CollSubspaceFone)\} \ ,
\end{equation*}
or equivalently as:
\begin{equation*}
  \dH(\CollSubspaceFone,\CollSubspaceFtwo) := \max \left( \sup_{\subspaceFone \in \CollSubspaceFone} \inf_{\subspaceFtwo \in \CollSubspaceFtwo} \dist(\subspaceFone,\subspaceFtwo), \sup_{\subspaceFtwo \in \CollSubspaceFtwo}\inf_{\subspaceFone\in\CollSubspaceFone} \dist(\subspaceFone,\subspaceFtwo) \right) \ .
\end{equation*}
\end{definition}

\begin{proposition}\textnormal{\textbf{\citep{buragobook}}} Consider the metric space $\multivarspace$, then:
\begin{enumerate}
\item $\dH$ is a semi-metric on $2^\multivarspace$ (the set of all subsets of $\multivarspace$) ,
\item $\dH(\CollSubspaceFone,\hat{\CollSubspaceFone})=0$ for any $\CollSubspaceFone \subset \multivarspace$ where $\hat{\CollSubspaceFone}$ denotes the closure of $\CollSubspaceFone$ ,
\item If $\CollSubspaceFone$ and $\CollSubspaceFtwo$ are closed subsets of $\multivarspace$ and $\dH(\CollSubspaceFone,\CollSubspaceFtwo)=0$, then $\CollSubspaceFone=\CollSubspaceFtwo$.
\end{enumerate}
\end{proposition}

Then it turns that the Hausdorff distance  is a true metric \emph{iff} the considered sets are closed.
A known limitation is that the Hausdorff distance is non-smooth.

This distance could be reformulated in term of a {\it correspondence}.
This allows to explicit the link between the Hausdorff distance and the class of Wasserstein distances that will be introduced in the sequel (cf. Section~\ref{sec:wasserstein}). 
% \begin{definition}
%   Let $\CollSubspaceFone$ and $\subspaceFtwo$ be two sets, a subset $R \subset \CollSubspaceFone \times \subspaceFtwo$ is a \emph{correspondence} if $\forall \subspaceFone \in \CollSubspaceFone$ there is $\subspaceFtwo \in \subspaceFtwo$ s.t. $(\subspaceFone,\subspaceFtwo)\in R$ and $\forall \subspaceFtwo \in \subspaceFtwo$ there is $\subspaceFone \in \CollSubspaceFone$ s.t. $(\subspaceFone,\subspaceFtwo)\in R$.
% \end{definition}
% \quentin{ Il doit y avoir une erreur dans cette Def. Perso, je mettrai:
\begin{definition}
  Let $\CollSubspaceFone$ and $\CollSubspaceFtwo$ be two sets, a subset $R \subset \CollSubspaceFone \times \CollSubspaceFtwo$ is a \emph{correspondence} if $\forall \subspaceFone \in \CollSubspaceFone$ there is $\subspaceFtwo \in \CollSubspaceFtwo$ s.t. $(\subspaceFone,\subspaceFtwo)\in R$ and $\forall \subspaceFtwo \in \CollSubspaceFtwo$ there is $\subspaceFone \in \CollSubspaceFone$ s.t. $(\subspaceFone,\subspaceFtwo)\in R$.
\end{definition} 
%}

Let $\mc{R}(\CollSubspaceFone,\CollSubspaceFtwo)$ denotes the set of all possible correspondences between $\CollSubspaceFone$ and $\CollSubspaceFtwo$.
It is possible to distinguish between a correspondence $\phi$ between $\CollSubspaceFone$ and $\CollSubspaceFtwo$ defined on the subset $R \subset \CollSubspaceFone \times \CollSubspaceFtwo$ as
\begin{displaymath}
  R = \{ (\subspaceFone,\subspaceFtwo) \in \CollSubspaceFone \times \CollSubspaceFtwo : \subspaceFtwo \in \phi(\subspaceFone), \subspaceFone \in \phi(\subspaceFtwo)\}
\end{displaymath}
and the subset $R$, which is also called the graph of $\phi$.

\begin{proposition}
Let $\CollSubspaceFone=\{\subspaceFone_i\}_{i \in I}$ and $\CollSubspaceFtwo=\{\subspaceFtwo_j\}_{j\in J}$ be two subsets of the metric space $(\multivarspace,d)$. The Hausdorff distance between $\CollSubspaceFone$ and $\CollSubspaceFtwo$ could be rewritten as:
\begin{equation*}
  \label{eq:haus2}
  \dH(\CollSubspaceFone, \CollSubspaceFtwo) = \inf \{ \sup_{\subspaceFone,\subspaceFtwo \in R} d(\subspaceFone, \subspaceFtwo) : R \in \mc{R}(\CollSubspaceFone,\CollSubspaceFtwo)\} \ ,
\end{equation*}
or equivalently
\begin{equation}
  \label{eq:haus2alt}
  \dH(\CollSubspaceFone, \CollSubspaceFtwo) = \inf_{R \in \mc{R}(\CollSubspaceFone,\CollSubspaceFtwo)} \sup_{\subspaceFone,\subspaceFtwo \in R} d(\subspaceFone, \subspaceFtwo) \ .
\end{equation}
\end{proposition}

With  the view of this last proposition, one can easily construct new set-metrics by replacing the $\sup$ operator in Equation~\eqref{eq:haus2alt} by any $\ell_p$ norm, hence leading to  more interesting smooth set-metrics. Such metrics belong to the class of Wasserstein distances by restricting the Borel measure to the Dirac function. The next section clarifies this statement.

\subsection{Wasserstein Distance}
\label{sec:wasserstein}

Let us define by ${\cal C}(\multivarspace)$ any collection on $\multivarspace$ and by $\multivarspace_W = \{ (\subspaceFone,\measure_\subspaceFone) : \subspaceFone \in \multivarspace \}$ where $\measure_\subspaceFone$ is a Borel measure.% with $\support{\measure_\subspaceFone} = \subspaceFone$.

\begin{definition}
  Let $\CollSubspaceFone,\CollSubspaceFtwo \in \multivarspace_W$.
  A measure $\measure$ on the product space $\CollSubspaceFone \times \CollSubspaceFtwo$ is a coupling of $\measure_\CollSubspaceFone$ and $\measure_\CollSubspaceFtwo$ if
  \begin{equation}
    \label{eq:coupling}
    \measure(\bar{\CollSubspaceFone} \times \CollSubspaceFtwo) = \measure_\CollSubspaceFone(\bar{\CollSubspaceFone}), \quad \measure(\CollSubspaceFone \times \bar{\CollSubspaceFtwo}) = \measure_\CollSubspaceFtwo(\bar{\CollSubspaceFtwo})
  \end{equation}
  for all Borel sets $\bar{\CollSubspaceFone}\subset \CollSubspaceFone,\;\bar{\CollSubspaceFtwo}\subset \CollSubspaceFtwo$.
\end{definition}
We denote by $\mc{M}(\measure_\CollSubspaceFone,\measure_\CollSubspaceFtwo)$ the set of all couplings of $\measure_\CollSubspaceFone$ and $\measure_\CollSubspaceFtwo$.
An interesting property is that given $\measure\in\mc{M}(\measure_\CollSubspaceFone,\measure_\CollSubspaceFtwo)$ then $R(\measure) = \support{\measure}$ is in $\mc{R}(\CollSubspaceFone,\CollSubspaceFtwo)$.

\begin{definition}
Let  $\CollSubspaceFone,\,\CollSubspaceFtwo\in\multivarspace_W$, and $\measure$ a coupling as defined in Equation~\eqref{eq:coupling}. 
Given $p>0$, the Wasserstein distance  is defined as:
\begin{align}
  \label{eq:wasserstein1}
  \dW^p(\CollSubspaceFone,\CollSubspaceFtwo)=&\inf_{\measure\in\mc{M}(\measure_\CollSubspaceFone,\measure_\CollSubspaceFtwo)}\left( \int_{\CollSubspaceFone \times \CollSubspaceFtwo} d(\subspaceFone,\subspaceFtwo)^p d\measure(\subspaceFone,\subspaceFtwo) \right)^{\frac{1}{p}}&~\text{if~~}  1 \leq p\\
  \label{eq:wasserstein2}
  \dW^p(\CollSubspaceFone,\CollSubspaceFtwo)=&\inf_{\measure\in\mc{M}(\measure_\CollSubspaceFone,\measure_\CollSubspaceFtwo)} \int_{\CollSubspaceFone \times \CollSubspaceFtwo} d(\subspaceFone,\subspaceFtwo)^p d\measure(\subspaceFone,\subspaceFtwo) &~\text{if~}  0 < p <1\\
  \label{eq:wasserstein3}
  \dW^\infty(\CollSubspaceFone,\CollSubspaceFtwo)=&\inf_{\measure\in\mc{M}(\measure_\CollSubspaceFone,\measure_\CollSubspaceFtwo)}\sup_{(\subspaceFone,\subspaceFtwo)\in\mc{R}(\measure)}d(\subspaceFone,\subspaceFtwo)  &\text{~otherwise} \ .
\end{align}
\end{definition}
This distance is also called the Wasserstein-Kantorovich-Rubinstein.

\begin{proposition}
Let $\CollSubspaceFone,\CollSubspaceFtwo$ be two subsets of $\multivarspace$, if we consider as a measure $\measure$ the Dirac one, then the Wasserstein distance is related to the Hausdorff distance by the following inequality:
\begin{equation}
  \label{eq:hausdwasser}
  \dH(\CollSubspaceFone,\CollSubspaceFtwo) \leqslant \dW^\infty\left( (\CollSubspaceFone,\measure_\CollSubspaceFone),(\CollSubspaceFtwo,\measure_\CollSubspaceFtwo)\right) \ .
\end{equation}
%for all $\measure_\CollSubspaceFone$ and $\measure_\CollSubspaceFtwo$ such that $\CollSubspaceFone = \support{\measure_\CollSubspaceFone}$ and $\CollSubspaceFtwo = \support{\measure_\CollSubspaceFtwo}$ or in the case of probability measure $\measure_\CollSubspaceFone(x)=\int_y\delta_y(x)dy$
\end{proposition}

\subsection{Set-Metrics for Frames and dictionaries}
\label{sec:met4frames}

Metrics in Equations~\eqref{eq:wasserstein1}, \eqref{eq:wasserstein2}, \eqref{eq:wasserstein3} and \eqref{eq:hausdwasser} are defined on Grassmannian space.
As indicated in Section~\ref{sec:prelimMet}, $(\multivarspace, \dist)$ is a separable metric space allowing to compute a distance between the collection of subspaces spanned by a frame and the collection of subspaces spanned by another.
The distance between two frames in the frame space, that is $\vecspace$, is defined as follows.

\begin{definition}
Let $\CollFone=\{\Fone_i\}_{i\in I}$ and $\CollFtwo=\{ \Ftwo_j \}_{j\in J}$ be two frames of the $\N \times \Vdim$ vector space $\vecspace$.
We define a distance between these two frames as:
\begin{equation}
  \label{eq:metric}
  \dF(\CollFone,\CollFtwo) = d_S (\CollSubspaceFone,\CollSubspaceFtwo) \ ,
\end{equation}
where $\CollSubspaceFone=\{\subspaceFone_i : \subspaceFone_i = \tmop{span}(\Fone_i), i\in I \}$ and $\CollSubspaceFtwo=\{\subspaceFtwo_j : \tmop{span}(\Ftwo_j), j\in J\}$, and $d_S$ is either $\dH$ or $\dW$. 
\end{definition}

From this definition, we can formulate the proposition:
\begin{proposition}
  Let $\mc{C}(\vecspace)$ be any collection on $\vecspace$.% and let $\mc{C}_W := \{ (\Fone, \pi_{\Fone}) : \Fone \in \vecspace \}$, where $\pi_{\Fone}$ is a Borel measure.
  Then the following holds:
  \begin{itemize}
  \item   $\dF$ is pseudo-metric and hence $(\mc{C}(\vecspace), \dF)$ is pseudo-metric space,
  \item $\dF$ is invariant by linear combinations.
  \end{itemize}
  % $\mc{C}(\vecspace)$ together with the distance $\dF$ define a pseudo-metric measurable space  and is invariant by linear combination.
%  The distance defined in Equation~\eqref{eq:metric} is a pseudo-metric and is invariant by linear combination.
\end{proposition}

\begin{proof}
  The proof is a direct consequence of Equation~\eqref{eq:metric} since $\dF$ is defined as a distance between subspaces.
\end{proof}

The frame distance $\dF$ is defined as the distance between their subspaces, that is $\dF$ is acting in the frame space $\vecspace$ whereas the distance $d_S$ is acting in the Grassmannian space $\multivarspace$.
As an element $\subspaceFone$ of $\multivarspace$ is a subspace, there exist an infinite number of elements $\Fone$ in $\vecspace$ spanning $\subspaceFone$.
Thus two distinct frames $\CollFone_1 \neq \CollFone_2$, that is two collections $\{\Fone_i^1\}_{i\in I}$ and $\{\Fone_i^2\}_{i\in I}$ of elements in $\vecspace$, could span the same collection of subspaces $\CollSubspaceFone=\{\subspaceFone_i\}_{i\in I}$ in $\multivarspace$.
In other words, a distance $\dF(\CollFone_1,\CollFone_2)=d_S(\CollSubspaceFone,\CollSubspaceFone)=0$ could exist for two separate frames $\CollFone_1\neq\CollFone_2$.
As the separability axiom does not hold, $\dF$ is a pseudo-metric and the separability axiom is relaxed to the identity axiom: $d(x,x)=0$, $\forall x \in X$.

Pseudo-metric spaces are similar to metric spaces and important properties of metric spaces could be proven as well for pseudo-metric spaces, so the fact that $(\mc{C}(\vecspace), \dF)$ is a pseudo-metric space has a very limited impact for further theoretical developments.
This situation is even desirable, as the distance between frames is defined through their associated subspaces yields the invariance by linear combinations, which is an essential property for comparing multivariate dictionaries.

The proposition of set-metrics for frames is a major contribution for the dictionary learning community as it is the first proposition to define a real metric in the dictionary space.
Previous works rely on heuristics to assess differences between dictionaries and no tractable framework has been proposed to compare dictionaries.
The work of \citet{SKR11} is a noticeable attempt to quantify the similarity between two dictionaries, without achieving to define a metric.

The proposed pseudo-metrics -- relying on Hausdorff and Wasserstein set-metrics, as well as chordal, Fubiny-Study or other ground metrics -- have been described in the literature and are easy to implement.
The following section described a multivariate dictionary learning algorithm which is then applied in Section~\ref{sec:simu_data} on synthetic data and in Section~\ref{sec:reseeg} on real world signals, demonstrating that these metrics could be applied ``out-of-the-box'' to design better dictionary learning algorithms or to assess machine learning datasets.

%$(\multivarspace,\delta)$ is a metric space for the multivariate dictionaries. Each point of this space is a frame (i.e. multivariate dictionary) $\mathcal{F}=\{F_m\}_{m=1}^M$, with atoms $F_m \in \mathbb{R}^{N \times V}$.
% $\mathbb{F}=\{F_i\}$ and $\mathcal{F} \subseteq \mathbb{F}$, thus $\mathcal{F} \in \mathcal{P}(F)$. $\mathcal{F}$ are Grassmanian manifolds $G(V,N)$.

% The Wasserstein distance is define as:
% \begin{displaymath}
%   d_w^p(\mathcal{F}_1,\mathcal{F}_1) = \min_{\mu \in \mathcal{M}(\mathcal{F}_1,\mathcal{F}_2)} \left( \int_{\mathcal{F}_1 \times \mathcal{F}_2}\delta^p (F_1,F_2)d\mu(F_1,F_2)\right)^{\frac{1}{p}}
% \end{displaymath}
% with $d^p_w:\mathcal{P}(X) \rightarrow \mathbb{R}$

%zz: \textcolor{red}{Introduire EMD comme un cas particulier.}

%%%%%%%%%%%%%%%%%%%%%%%%%%%%%%%%%%%%%%%%%%%%%%%%%%%%%%%%%%%%%%%%%%%

\section{Dictionary Learning for Multivariate Signals}
\label{sec:multivardla}
%\sylvain{Dictionaries are noted $\dico$ instead of $\boldsymbol\Phi$}

Hereafter, to illustrate the metrics between frames with concrete examples, we will considered a multivariate dictionary $\Dico$ as a collection of $\M$ multivariate atoms $\dico_\find \in \Re^{\N \times \Vdim}$. In this section, models and dictionary learning algorithms for multicomponent signals are detailed.

\subsection{Model for Multivariate Dictionaries}

We consider a multivariate signal $\data \in \mathbb{R}^{\N \times \Vdim}$, whose decomposition is carried out on the multivariate dictionary $\Dico$ thanks to the multivariate model~\citep{Barthelemy2012}:
\begin{equation}
	\label{eq:decomp_multivar} 
	\data = \sum_{\find=1}^\M \coef_\find \: \dico_\find + E \ ,
\end{equation} 
where $\coef_\find \in \Re$ denotes the coding coefficient and $E \in \Re^{\N \times \Vdim}$ the residual error.
Two problems need to be handled on this model: the first one, called sparse approximation, estimates the coefficient vector $\coef$ when the dictionary $\Dico$ is fixed, and the second one, called dictionary learning, estimates the best dictionary $\Dico$ from the dataset $\Data=\left\{ \data_\qind \right\}_{\qind=1}^\Q$. 

% \subsection{Multivariate sparse approximation}

Since the dictionary is redundant, $\M \gg \N$, the linear system of \eqref{eq:decomp_multivar} is thus under-determined and has multiple solutions. 
The introduction of constraints such as sparsity allows to regularize the solution.
With the sparse approximation, only $K$ active atoms are selected among the $\M$ possible ones and the associated coefficients vector $\coef$ is computed to maximize the approximation of the signal $\data$. 
The multivariate sparse approximation can be formalized as:
\begin{equation}
	\label{eq:sparse_app}
	\text{min}_\coef \left\| \: \data - \sum_{\find=1}^\M \coef_\find \: \dico_\find \: \right\|^2_F \ \text{s.t.} \ \left\| \coef \right\|_0  \leqslant  K \ ,	
\end{equation}
where $K\! \ll \!\M$ is a constant. But this problem is NP-hard \citep{Davis1994a}.
In the univariate case, non-convex pursuits tackle it sequentially, such as matching pursuit (MP) \citep{Mallat1993a} or orthogonal matching pursuit (OMP) \citep{Pati1993}. 
Many other sparse approximation algorithms are reviewed in \citet{Tropp2010}. 
In the multivariate case, Equation~\eqref{eq:sparse_app} is solved by the multivariate OMP (M-OMP) \citep{Barthelemy2012}.

%\subsection{Multivariate Dictionary Learning}

Dictionary learning presented in Section~\ref{sec:dlgrass} is now extended to the multivariate case.
Adding an index $\qind$ to variables to denote their link to the signal $\data_\qind$, the multivariate dictionary learning problem is formalized by:
\begin{equation}
	\label{eq:dl1}
   \begin{split}
     \text{min}_{\Dico} \ \sum_{\qind=1}^\Q \ \text{min}_{\coef_\qind} \ \left\| \: \data_\qind - \sum_{\find=1}^\M \coef_{\find,\qind} \: \dico_\find \: \right\|^2_F \text{s.t. } 
      & \left\| \coef_\qind \right\|_0  \leqslant  K,\, \qind=1 \ldots \Q
     \\
     \ \text{and } & 
     \left\| \dico_\find \right\|_F = 1,\, \find=1\ldots \M \ .
   \end{split}
\end{equation}
The multivariate DLA (M-DLA) proposed in \citet{Barthelemy2013a} solves this problem by learning a multivariate dictionary from the multivariate dataset $\Data$.
This algorithm has been used to study multivariate time-series such as 2D spatial trajectories \citep{Barthelemy2012} and EEG signals \citep{Barthelemy2013a}.

\subsection{Model for Multicomponent Dictionaries}

Some other works use multicomponent dictionaries to analyze multicomponent data: audio-visual data \citep{MON07, Monaci2009}, electro-cardiogram signals \citep{Mailhe2009a}, color images \citep{Mairal2008}, hyperspectral images \citep{MOU09}, stereo images \citep{Tosic2011a}.
But their decomposition model is different from the one used with multivariate dictionaries.

With multicomponent dictionaries, each component is associated with a univariate atom multiplied by its own coefficient, contrary to the multivariate model of Equation~\eqref{eq:decomp_multivar} where components form a multivariate atom multiplied by one unique coefficient, identical for all components.
Let recall that $\data=[y_1, \ldots, y_\Vdim]$ is a $\N \times \Vdim$ matrix and $\dico_\find = [\fone_{\find,1},\ldots,\fone_{\find, \Vdim}]$ is a $\N \times \Vdim$ matrix.
In this paragraph, the coefficients $\coef_\find=[\coef_{\find,1},\ldots,\coef_{\find, \Vdim}]$ form a $\Vdim$-dimensional vector.
%For this paragraph, we precise that $\data = \left\{ \: y_k \in \mathbb{R}^{\N} \: \right\}_{k=1}^\Vdim$, $\dico_\find = \left\{ \: \fone_{\find,k} \in \mathbb{R}^{\N} \: \right\}_{k=1}^\Vdim$, and $\coef_\find = \left\{ \: \coef_{\find,k} \in \mathbb{R} \: \right\}_{k=1}^\Vdim$.
Each component $k$ is decomposed independently:
\begin{equation}
\label{eq:mcompdico}
	\y_k = \sum_{\find=1}^\M \coef_{\find,k} \: \fone_{\find,k} + e_k \ , \: k=1\ldots \Vdim \ ,
\end{equation} 
thus the coefficients $\coef$ is a $\M \times \Vdim$ matrix.
To compare the multivariate model of Equation~\eqref{eq:decomp_multivar} with the multicomponent model of Equation~\eqref{eq:mcompdico}, the latter can be reformulated as:
\begin{equation*}
	\data = \sum_{\find=1}^\M \coef_\find \otimes \dico_\find + E \ ,
\end{equation*}
%with the coefficients $\coef_\find \in \mathbb{R}^{\Vdim}$ and 
where $\otimes$ is defined as the element-wise product along dimension $\Vdim$.
Thus, in the multicomponent approach, multivariate signals are transformed in parallel univariate signals, causing an important information loss.

Sparse approximations adapted to this model estimate the multicomponent coefficients associated to the selected active atoms.
The relation between the components of an atom is only considered during the atom selection: the index $\find$ is the same for the $\Vdim$ components $\fone_{\find,k}$. The selected atom $\dico_{\find}$ is thus jointly chosen between components thanks to a sparse mixed norm \citep{Gribonval2007, Rakotomamonjy2011}.
Algorithms adapted to multicomponent approach are not detailed more here since this model is not used in the sequel.

\subsection{Model for Rotation Invariance}
\label{sec:modelrotinv}

The multivariate model described in Equation~\eqref{eq:decomp_multivar} is now extended to include rotation invariance. A rotation matrix $R_\find \in \Re^{\Vdim \times \Vdim}$ is added to each atom $\dico_\find \in \Re^{\N \times \Vdim}$, with $R_\find R_\find^T = I_\Vdim$. Thus, the decomposition model becomes:
\begin{equation}
	\data = \sum_{\find=1}^\M \coef_\find \: \dico_\find \: R_\find + E  \ .
	\label{eq:decomp_ndri} 
\end{equation} 
This model is called \emph{nD Rotation Invariant} (nDRI) in \citet{Barthelemy2013}, since multivariate atoms $\dico_\find$ are now able to rotate.

The multivariate sparse approximation described in Equation~(\ref{eq:sparse_app}) is extended to this rotation invariant model.
Coefficients $\coef=\left\{ \coef_\find \right\}_{\find=1}^\M$ and rotation matrices $\boldsymbol{R}=\left\{ R_\find \right\}_{\find=1}^\M$ are estimated with the nDRI-OMP
\citep{Barthelemy2013} such as: 
\begin{equation*}
	\begin{split}
	\text{min}_{\coef,\boldsymbol{R}}  \left\| \: \data - \sum_{\find=1}^\M \coef_\find \: \dico_\find \: R_\find \: \right\|^2_F \ \text{s.t.} 
	& \ \left\| \coef \right\|_0  \leq  K
	\\
	\ \text{and }
	& R_\find R_\find^T = I_\Vdim, \, \find=1\ldots \M \ .	
	\end{split}
\end{equation*}
The core of the nDRI-OMP is the orthogonal Procrustes problem. 
The goal is to estimate the coefficient $\alpha \in \mathbb{R}$ and the rotation matrix $R \in \Re^{\Vdim \times \Vdim}$ which optimally register the atom $\dico \in \Re^{\N \times \Vdim}$ on the signal $Z \in \Re^{\N \times \Vdim}$:
\begin{equation*}
	(\tilde{\alpha},\tilde{R}) = \text{arg} \: \text{min}_{\alpha,R}  \left\| \: Z - \alpha \: \dico \: R \: \right\|^2_F \ \ \text{s.t.} \ \ R R^T = I_\Vdim \ .	
\end{equation*}
The function which computes this step in denoted $(\tilde{\alpha},\tilde{R}) = \text{nD\_Registration}(\dico,Z)$ in the following.

The multivariate dictionary learning described in Equation~(\ref{eq:dl1}) is extended to the rotation invariant case as:
\begin{equation*}
   \begin{split}
     \text{min}_{\Dico} \ \sum_{\qind=1}^\Q \ \text{min}_{\coef_\qind,\boldsymbol{R}_\qind} \ \left\| \: \data_\qind - \sum_{\find=1}^\M \coef_{\find,\qind} \: \dico_\find \: R_{\find,\qind} \: \right\|^2_F  
     \\ \text{s.t. } \left\| \coef_\qind \right\|_0  \leqslant  K, \qind=1 \ldots \Q
     \\
     \ \text{and }  R_{\find,\qind} R_{\find,\qind}^T = I_\Vdim, \, \find=1\ldots \M , \, \qind=1 \ldots \Q
     \\ 
     \ \text{and } \left\| \dico_\find \right\|_F = 1,\, \find=1\ldots \M \ . 
   \end{split}
\end{equation*}
The algorithm which solves this problem is called nDRI-DLA \citep{Barthelemy2013} and it has been applied in a trivariate case to study 3D spatial trajectories.

%%%%%%%%%%%%%%%%%%%%%%%%%%%%%%%%%%%%%%%%%%%%%%%%%%%%%%%%%%%%%%%%%%%

% \section{Detection rates and distances comparison on synthetic data}

\section{Experiments on Synthetic Data}
\label{sec:simu_data}

%In this section, the introduced methods are applied to simulation data, and compared to evaluate their performance.

This work is the first attempt to assess qualitatively or quantitatively the dictionary learning algorithms with real metrics.
Metrics on frames have several immediate and useful applications to improve our understanding of learning algorithms and datasets in a dictionary-based framework.
This section is devoted to show why relying on metrics allows to dramatically improve the assessment of dictionary learning algorithms.
More precisely, a set of experiments is conducted in Section~\ref{sec:ressyn} to reproduce state-of-the-art results on synthetic dataset and shows how the different proposed metrics compared with the commonly used indicators.
% More precisely, the improvements offered by metrics is shown for a common evaluation setup used in the DLA community: recovering a generating dictionary from noisy synthetic inputs. % detection rates on synthetic data.
A second set of experiments, presented in Section~\ref{sec:reseeg}, shows how metrics could be used on multivariate signal, in the context of Brain-Computer Interface, to evaluate qualitatively datasets of brain signal.
%The set-metrics allows to effortlessly adapt clustering algorithms and lead to interesting conclusion on the dataset of BCI Competition IV.

% The next section will show how metrics could be used on real multivariate signal, in the context of Brain-Computer Interface, to evaluate qualitatively datasets of brain signal.
% Two sets of experiments are conducted, the first one reproduces state-of-the-art results and shows how the different proposed metrics compared with the commonly used detection rate. %are related
% The learning rate and the robustness to noise are described in the Section~\ref{sec:ressyn}.
% In the second set of experiments, the metrics are applied on dictionary learned over real EEG signal for Brain-Computer Interface system.
% The set-metrics allows to effortlessly adapt clustering algorithms and lead to interesting conclusion on the dataset of BCI Competition IV.
% Those results are presented in Section~\ref{sec:reseeg}.
% Dictionary learning algorithms are classically compared with detection rates. In this section, we will show the advantage of distances for such task. \textcolor{red}{intro a reprendre.}

% In the following experimental results, we rely on the Wasserstein distance with $p=1$, which is sometimes called Kantorovich distance or also commonly known as Earth Mover's Distance (EMD).

\subsection{Experimental Protocol}
\label{sec:expproto}

%In this protocol, a fixed number of chosen signals, say $\M$, are used as a primitive set, a ground truth. 
The experimental protocol described hereafter was inspired by \citet{Aharon2006b}, \citet{Barthelemy2012} and \citet{Barthelemy2013}\footnote{In the context of rotation invariance, this experiment can be viewed as the extension of \citet{Barthelemy2013} in dimension $\Vdim=10$.}.
It is presented leaving out the explanations relative to the shift-invariant signal processing to lighten the description.

A dictionary with $\M$ multivariate atoms of $\Vdim=10$ components, generated randomly%or from structured signals
, serve as a generating synthetic data and is thus called the \emph{original dictionary}.
By combining a small number of atoms of the original dictionary, a training dataset ($\datastr$) is produced.
This training dataset is used to extract a \emph{learned dictionary} with at least $\M$ atoms.
The quality of the DLA is then evaluated by comparing how many atoms from the original dictionary have been recovered in the learned dictionary.
% This experiment was designed to test the recovery ability of the M-DLA and the nDRI-DLA.
% The experimental protocol described hereafter was inspired by \citep{Aharon2006b}, \citep{Barthelemy2012} and \citep{Barthelemy2013} \footnote{The following experiment is not presented in a shift-invariant case in order to lighten the experiment description.}.

Here, the original dictionary $\Dico$ of $\M=135$ normalized multivariate atoms is created from white uniform noise and the length of the atoms is $\N=20$ samples. 
The training set $\datastr$ is composed of $\Q=2000$ signals of length $\N$ and it is synthetically generated from dictionary $\Dico$.
Each training signal is generated as of the sum of three atoms, the coefficients and the atom indices being randomly drawn.
At first, no noise is added to the generated signals for the set of experiments conducted in Section~\ref{sec:ressyn}. 
An analysis of the noise sensitivity is proposed in Section~\ref{sec:noisesensitivity}.
The dictionary initialization is made on the training set, and the learned dictionary $\hat{\Dico}$ is returned after $80$ iterations over the training set.

The experimental protocol was slightly changed to give the training set $\datarot$. 
In this case, a rotation is applied for each atoms composing the training signals.
The coefficient of the rotation matrix, see Equation~\eqref{eq:decomp_ndri}, are drawn randomly. % with the multivariate atoms, i.e. $R_\Mind = Id$.

\subsection{Detection Rates and Metrics Calculus}

In the experiments, a learned atom $\hat{\dico}_\find$ is considered as detected, or recovered, if its correlation value $\nu_\find$ with its corresponding original atom $\dico_\find$ is greater than a chosen threshold $s$.
In the multivariate case of M-DLA described in Equation~\eqref{eq:decomp_multivar}, it is expressed as:
\begin{equation}
	\label{eq:detection_cond_2}
	\nu_\find = \left| \left\langle \dico_\find,\hat{\dico}_\find \right\rangle \right| \geqslant s \ .
\end{equation}
where $s$ is a threshold fixed at $0.99$ or $0.97$. The \emph{detection rate} is defined as the percentage of the original dictionary atoms which are recovered in the learned dictionary and it is denoted as $t_{0.99}$ and $t_{0.97}$.
This threshold-based approach is very common in the DLA community to obtain a quantitative measure on dictionaries, but it is not a metric.
Remark that this detection rate is invariant to sign: $\hat{\dico}_\find$ is considered recovered if it is close to the original atom $\dico_\find$ or to its opposite $-\dico_\find$. The invariance to scale is already obtained since atoms are normalized.

This is naturally extended to include rotation invariance, as shown in Equation~\eqref{eq:decomp_ndri} for nDRI-DLA setup, by computing the correlation $\nu_\find$ between $\dico_\find$ and $\hat{\dico}_\find$ after nD registration:
\begin{equation}
	\label{eq:detection_cond}
	\nu_\find \geqslant s \ \ \text{with} \ \ (\nu_\find,\cdot)=\text{nD\_Registration}(\hat{\dico}_\find,\dico_\find) \ .
\end{equation}
Due to the nD registration step, this detection rate is invariant to rotation: $\hat{\dico}_\find$ is considered recovered if it is a rotated version of the original atom $\dico_\find$.

In the following, two multivariate DLA algorithms are tested: M-DLA and the rotation invariant nDRI-DLA.
The M-DLA is tested on the dataset $\datastr$, which does not include rotation of original atoms.
The nDRI-DLA is evaluated on the two datasets $\datastr$ %(denoted \emph{nDRI-DLA (a)} setup hereafter) 
and $\datarot$. %(\emph{nDRI-DLA (b)}).
The assessment of nDRI-DLA for both datasets relies on the detection rate based on Equation~\eqref{eq:detection_cond}.
The evaluation of the M-DLA detection rate is based on Equation~\eqref{eq:detection_cond_2}. 

% In the following, \textit{nDRI-DLA (a)} will denote the results for the training set $\mathcal{Y}_1$ and the detection condition of Equation~\eqref{eq:detection_cond}; \textit{nDRI-DLA (b)} will denote the results for the training set $\mathcal{Y}_2$ and the detection condition of Equation\eqref{eq:detection_cond}; and \textit{M-DLA} will denote the results for the training set $\mathcal{Y}_2$ and the detection condition of Equation \eqref{eq:detection_cond_2}.

Thanks to the Equation~\eqref{eq:metric}, it is possible to define real set-metric between dictionaries $\Dico$ and $\hat{\Dico}$.
With set-metrics, one important choice is the ground metric. 
For the multivariate dictionaries presented here, any distance described in Section~\ref{sec:du} could be selected as ground distance for Hausdorff or Wasserstein metric.
The ground metric allows to compute the distance between a learned atom and its associated atom in the original dictionary.
For now on, the notation is changed for the sake of clarity to indicate the set-metric as a subscript and the ground distance as a superscript, for example $\setmetric{W}{c}$ is the Wasserstein distance based on chordal ground distance.

To emphasize the importance of the ground distance choice and to clarify the results, only two ground distances are considered in this section.
The first ground metric is the chordal distance $d^{\tmop{c}}$, defined in Equation~\eqref{eq:chordaldef}, chosen for its invariance to linear combination, thus including rotations.
As all principal angle-based ground metric share these properties, the chordal have been selected only because it is a widely accepted choice for Grassmannian manifold.

The second ground metric is a simple Frobenius-based distance $d^{\,\tmop{f}}$, expressed for multivariate atoms $\dico_i$ and $\hat{\dico}_j$ as:
\begin{equation*}
  \label{eq:frodist}
%  d^{\tmop{f}}(A,B) = \norm{A-B}_F^2 = 2-2\tmop{trace}(A^TB)
  \left(d^{\,\tmop{f}}(\dico_i,\hat{\dico}_j)\right)^2 
  = \norm{ \hat{\dico}_j - \dico_i }_F^2 
  = 2 - 2\tmop{trace}( \hat{\dico}_j^T \dico_i  ) 
  = 2 \left( 1- \left\langle \dico_i,\hat{\dico}_j \right\rangle \right) \ ,
\end{equation*}
under the classical assumption that the dictionary atoms have unit norm in the Frobenius sense, $\norm{\dico_i}_F\,=\,\norm{\hat{\dico}_j}_F\,=1$.
This distance is a metric in Euclidean space but is not in Grassmannian spaces, it is thus not described in Section~\ref{sec:du}.
The distance $d^{\,\tmop{f}}$ is related to the detection rate of Equation~\eqref{eq:detection_cond_2}, but without the sign invariance: $\hat{\dico}_j$ is not considered recovered if it is close to $-\dico_i$. This distance is not invariant to linear transforms and hence to rotation and sign change.

The distances $d^{\,\tmop{c}}$ and $d^{\,\tmop{f}}$ served as ground distance for Hausdorff and Wasserstein set-metrics. For the latter, it is parametrized with $p=1$, see Equation~\eqref{eq:wasserstein1}, and also known as Earth Mover's distance or Kantorovich distance. Also, the measures are uniform on the whole support.
We investigate here the four possible combinations: Hausdorff over chordal ($\setmetric{H}{c}$) or over Frobenius-based distance ($\setmetric{H}{f}$) and Wasserstein over chordal ($\setmetric{W}{c}$) or Frobenius ($\setmetric{W}{f}$). The set-metric distance are rescaled to be visually comparable to detection rates which are given in percentage:
\begin{equation*}
  \begin{split}
	\tilde{\setmetric{\;\cdot}{c}} (\Dico,\hat{\Dico}) & = \frac{\sqrt{\Vdim}-\setmetric{\;\cdot}{c}(\Dico,\hat{\Dico})}{\sqrt{\Vdim}} \times 100 \, , 
   \\
  \tilde{\setmetric{\;\cdot}{f}} (\Dico,\hat{\Dico}) & = \frac{\sqrt{2}-\setmetric{\;\cdot}{f}(\Dico,\hat{\Dico})}{\sqrt{2}} \times 100 \ .    
  \end{split}
\end{equation*}

\subsection{Dictionary Recovering}
\label{sec:ressyn}

\begin{figure}[t!]
	\centering
   \pgfimage[interpolate=true,width=0.95\linewidth]{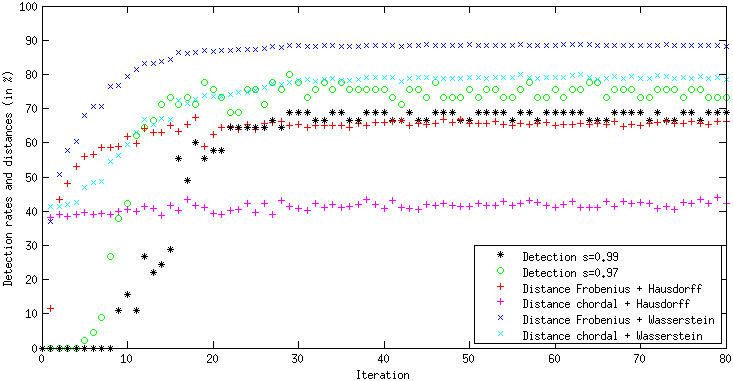}
	% \includegraphics[scale=0.5]{"Figures/AppSimu_M2"}
   % \resizebox{\linewidth}{!}{\input Figures/AppSimu_M2.tex}
	\caption{Evolution of detection rates and dictionary distances for M-DLA applied on $\datastr$ as a function of the learning iterations.}
	\label{fig:AppSimu_M}
\end{figure}

The M-DLA is evaluated on a dataset generated from a combination of the original atoms, that is dataset $\datastr$ which does not include rotations.
The results are presented on Figure~\ref{fig:AppSimu_M}, where the recovering rate of the learned dictionary, in term of distances or detection rates, is assessed for each iteration.

The detection rates $t_{0.99}$ or $t_{0.97}$ raises abruptly from 0 to around 70\% between iterations 5 and 20 and shows non-smooth variations around stable value, 67\% for $t_{0.99}$ and 72\% for $t_{0.97}$, for the rest of the iterations.
The Wasserstein set-metrics offer a better picture, this distance is more smooth and more stable that the detection rates.
This metric captures the fact that as the learned dictionary atoms are initialized with training signals of the dataset, the distance between the learned and the original dictionaries is not null, which is not the case for the detection rates.
When using the Wasserstein set-metrics, the choice of ground distance has only a limited influence: the Frobenius-based metric stabilizes earlier than the chordal-based one.
For the Hausdorff set-metrics, the experimental results confirm a known fact: the Hausdorff distances are less sensitive to internal variations inside the considered sets as they capture the distance of the most extreme point.
%Hausdorff distance are less sensitive of the internal variation as they capture the distance of the most extreme points.
Frobenius-based Hausdorff metric does not capture the changes caused by the learning algorithm, it stays with values around 40\% for all iterations, whereas the chordal-based one shows a similar pattern to the Wasserstein distances but with a smaller amplitude.

\begin{figure}[t!]
	\centering
   % \resizebox{\linewidth}{!}{\input Figures/AppSimu_nDRI_b2.tex}
   \pgfimage[interpolate=true,width=0.95\linewidth]{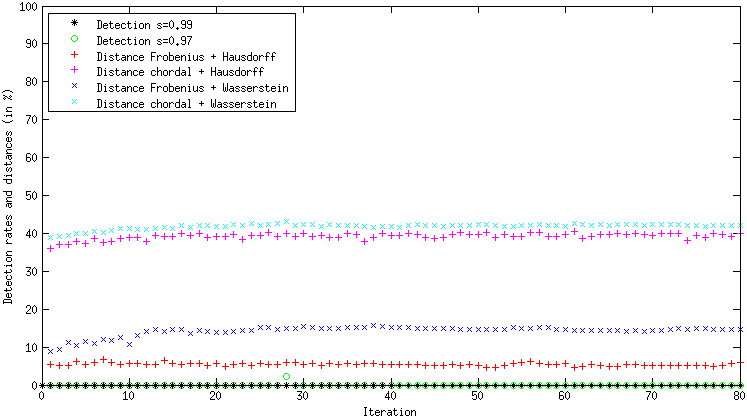}
	\caption{Same as Figure~\ref{fig:AppSimu_M} with nDRI-DLA on $\datastr$: it is a multivariate dictionary learning with rotation invariance (nDRI-DLA) applied on a dataset without rotation of the signals ($\datastr$).}
	\label{fig:AppSimu_nDRI_b}
\end{figure}
\begin{figure}[ht!]
	\centering
   % \resizebox{\linewidth}{!}{\input Figures/AppSimu_nDRI_a2.tex}
   \pgfimage[interpolate=true,width=0.95\linewidth]{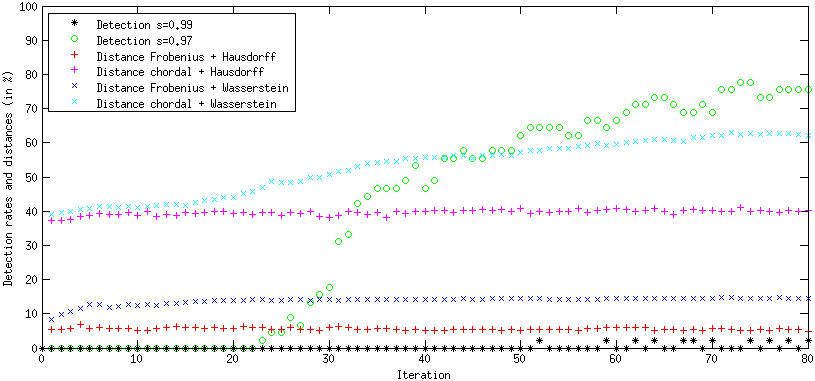}
	\caption{Same as Figure~\ref{fig:AppSimu_nDRI_b} with nDRI-DLA on $\datarot$: it is a multivariate dictionary learning with rotation invariance (nDRI-DLA) applied on a dataset with rotation of the signals ($\datarot$).}
	\label{fig:AppSimu_nDRI_a}
\end{figure}

% With nDRI-DLA, an interesting effect occur between $\datastr$ and $\datarot$.

The results obtained with nDRI-DLA on dataset $\datastr$ are shown on Figure~\ref{fig:AppSimu_nDRI_b}.
The set-metric -- Wasserstein on chordal and Frobenius or Hausdorff on chordal -- indicate that the algorithm slowly improve during the first 20 iterations before stabilizing.
As observed on M-DLA, the Hausdorff distance based on chordal shows no sensitivity to the dictionary evolution during the learning phase.
Another remark concerns the ground distance, here Frobenius or chordal, which has a major influence on the set-metric distance: the Wasserstein and Hausdorff distance based on chordal give very similar results.
The detection rate are null for the whole learning phase and thus completely uninformative.

The poor results of nDRI-DLA on $\datastr$ shown on Figure~\ref{fig:AppSimu_nDRI_b} are caused by the small size of the dataset given the large dimension of the multivariate components to learned: only $\Q=2000$ signals for learning $\M=135$ atoms with $\Vdim=10$ components.
This is enough to achieve decent performances with M-DLA, but as nDRI-DLA should learn the rotation matrix on top of recovering the atoms and their coefficient, nDRI-DLA failed to obtain correct performance on dataset $\datastr$.
However, with dataset $\datarot$ each input signal contains three atoms from the original dictionary with different rotation coefficient, it is thus easier for nDRI-DLA to recover the atoms and learn the rotation matrices, as it is demonstrated on Figure~\ref{fig:AppSimu_nDRI_a}.

With dataset $\datarot$, nDRI-DLA performances are shown on Figure~\ref{fig:AppSimu_nDRI_a}.
The detection rate with $t_{0.99}$ stays at 0, whereas $t_{0.97}$ abruptly increase from iteration 25 and displays large non-smooth fluctuations due to its binary nature.
Moreover, for $t_{0.99}$ very few atoms are considered as recovered, whereas for $t_{0.97}$ (which remains a selective rate) around 75\% of atoms are considered as recovered. 
This highlights the extreme sensitivity of the detection rate with respect to the threshold value $s$
The Frobenius distance is affected by rotations and thus not suited for this setup: neither Wasserstein nor Hausdorff set-metric based on Frobenius distance are able to capture the evolution of the learned dictionary atoms.
As explained previously the Frobenius-Wasserstein distance is not sensitive enough for the present setup.
The chordal-based Wasserstein gives useful information on the convergence of the algorithm, which is smoothly increasing at each iteration.
This example shows the importance of selecting an appropriate ground distance for set-metric, for example a rotation invariant distance such as the chordal is required to correctly assessed the nDRI-DLA with dataset $\datarot$.

\subsection{Noise Sensitivity Analysis}
\label{sec:noisesensitivity}

\begin{figure}[t!]
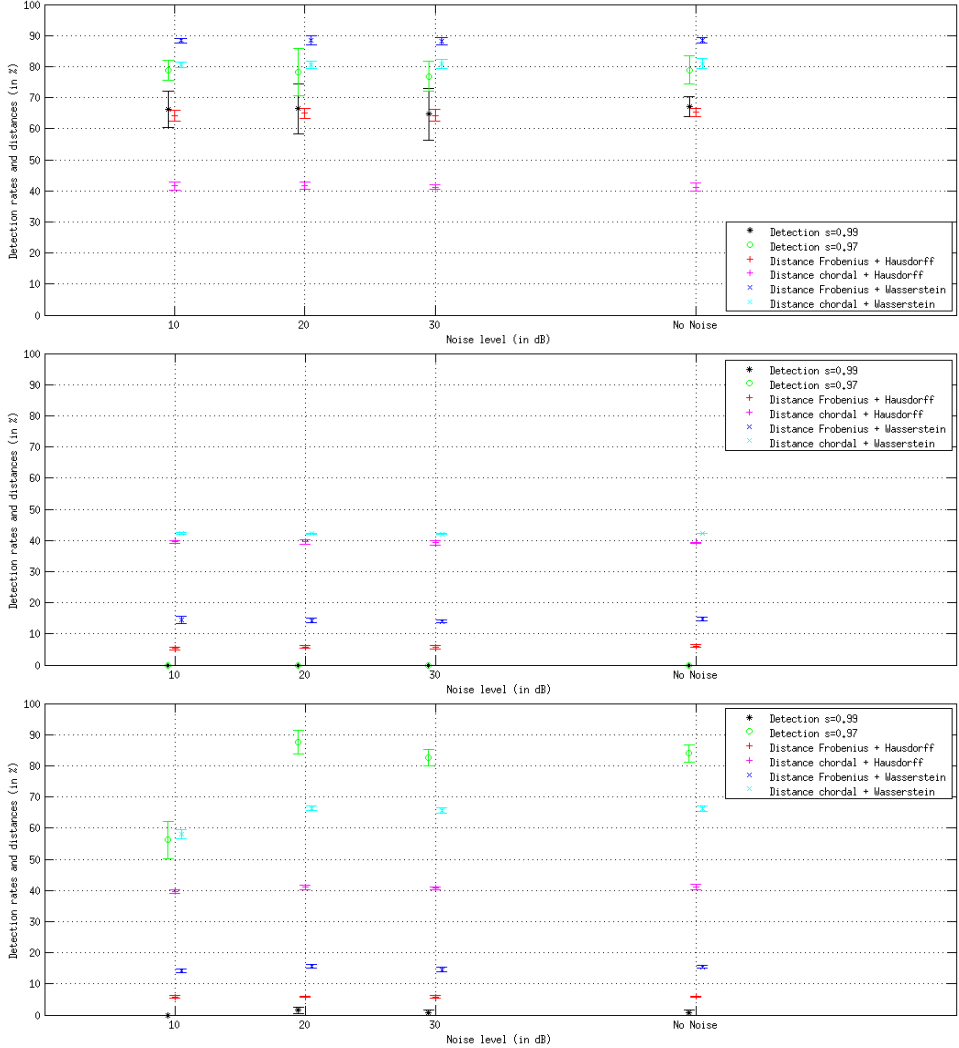

	\centering
   \pgfimage[interpolate=true,width=0.99\linewidth]{"AppSimu_M_ResFinal2"}
   \pgfimage[interpolate=true,width=0.99\linewidth]{"AppSimu_nDRI_b_ResFinal2"}
   \pgfimage[interpolate=true,width=0.99\linewidth]{"AppSimu_nDRI_a_ResFinal2"}
	\caption{Noise sensitivity analysis: averaged results for detection rates and distances with different noise level, that is 10, 20, 30 dB and no noise. (Top) M-DLA on dataset $\datastr$, (Center) nDRI-DLA on $\datastr$ (Bottom) nDRI-DLA on $\datarot$.}
	\label{fig:simnoise}
\end{figure}

For the next experiments, learning are repeated 10 times and Figure~\ref{fig:simnoise} shows the averaged detection rates and distances computed after 80 iterations.
The previous experiments are replicated for different noise levels in the three setups: M-DLA on $\datastr$, nDRI-DLA on $\datastr$ and nDRI-DLA on $\datarot$.
White Gaussian noise is added at several levels to obtain datasets with signal-to-noise ratio of $10$, $20$ and $30$ dB, and a dataset is kept without adding noise. 
On each plot, to ease the comparison, the last column indicates the results obtained in the noiseless setting after 80 iterations, as in Figures~\ref{fig:AppSimu_M}, \ref{fig:AppSimu_nDRI_b} and~\ref{fig:AppSimu_nDRI_a}. The results are also averaged over 10 repetitions for the noiseless case.

In the M-DLA case, on the left of Figure~\ref{fig:simnoise}, the detection rates and the distance are not affected by noise, showing the robustness of the learning algorithm on multivariate signals.
These conclusions could be extended to the nDRI-DLA case on dataset $\datastr$, at the center of Figure~\ref{fig:simnoise}.
The original dictionary is poorly recovered, less than half of the atoms are identified, but the results are not affected by the presence of noise, even at high level.
The detection rates $t_{0.97}$ and $t_{0.99}$ stay at 0 for all experiments and are thus completely inadequate for these evaluations.

Applying noise on the examples of dataset $\datarot$ as the same effect on nDRI-DLA than applying random rotations on each original atoms: the atoms ``pop out'' from the noise, that is the atoms are the only stable signals existing in the dataset, all other parameters vary.
It helps the algorithm to correctly reconstruct the original atoms, find their coefficient and their rotation matrix.
This explains that the dictionary reconstructing in the 20 and 30 dB cases have results as good as the noiseless case.
With only 10 dB, the noise yields small perturbations, corrupting the learned atoms and resulting in a decrease of recognition rate.

These experiments on synthetic dataset are aimed to demonstrate that direct applications of the previously described metrics offer immediate and handy tools for assessing dictionary learning algorithms.
To exploit the full potential of the proposed metrics, we have chosen to propose experiments on multivariate datasets, but all these results could be applied directly on univariate datasets.

%%%%%%%%%%%%%%%%%%%%%%%%%%%%%%%%%%%%%%%%%%%%%%%%%%%%%%%%%%%%%%%%%%%

\section{Experiments on EEG Signals}
\label{sec:reseeg}

\begin{figure}
  \centering
  % \resizebox{0.95\linewidth}{!}{\input exampleEEG2.tex}
  \pgfimage[interpolate=true,width=0.95\linewidth]{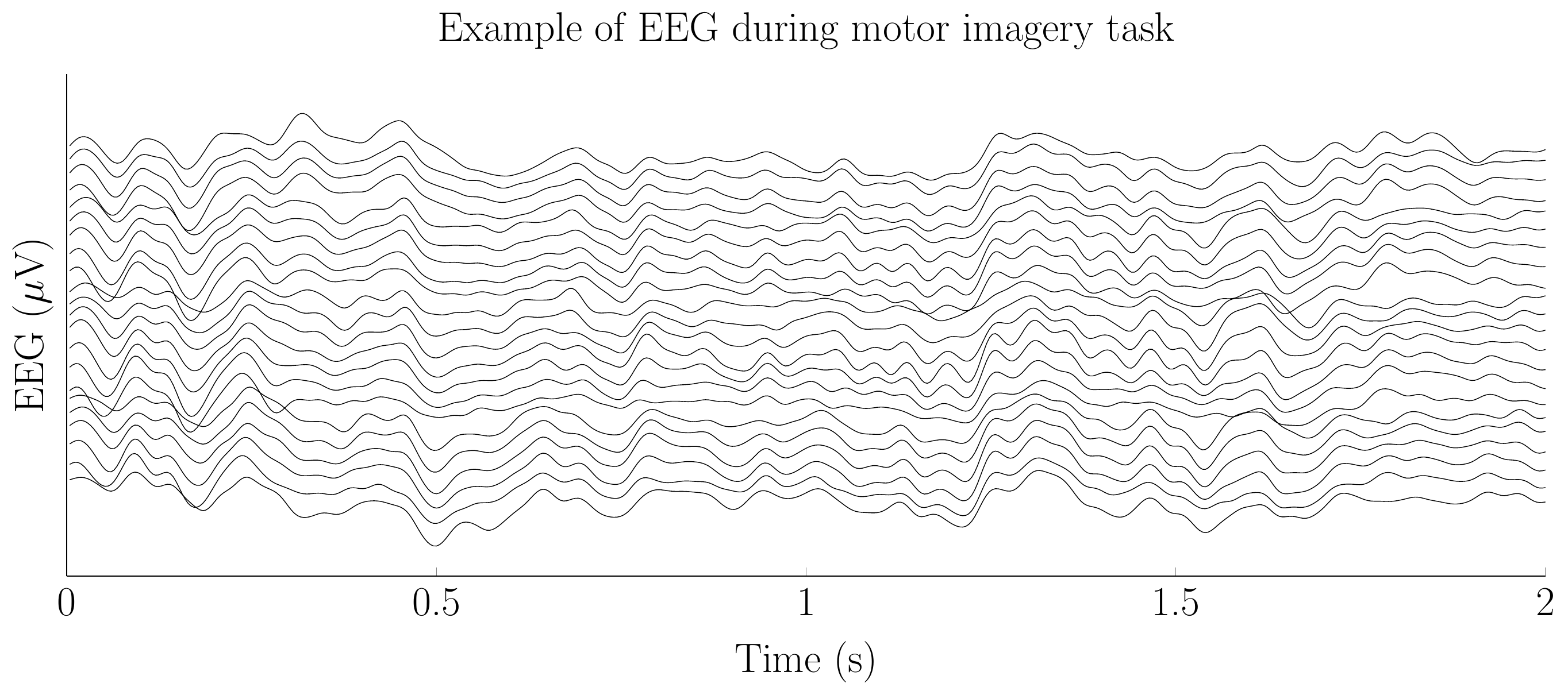}
  \caption{Example of an EEG during a 2 seconds motor imagery task. The electrical signals recorded from the 22 electrodes are plotted with a slight shift on the ordinate axis. Data from BCI Competition IV, set 2a, subject 7.}
  \label{fig:eegexample}
\end{figure}

This section demonstrates some possible applications implied by the definition of metrics between dictionaries.
We choose to present experiments on Brain-Computer Interface (BCI) because the spatio-temporal structure of the data offer, to some extend, an intuitive example of the application of multivariate set-metrics.
Most of the BCI systems rely on electroencephalographic (EEG) measurements, which are electric variations caused by brain waves are recorded on the surface of the head with several electrodes.
A dataset is thus an ensemble of timeseries, each of them representing the electric activity measured at a precise spatial location.
Due to signal propagation and to physical properties of the head, two neighboring spatial locations have highly correlated signals, as it can be seen on Figure~\ref{fig:eegexample}.

\subsection{Experimental Protocol}
\label{sec:proteeg}

Dictionary learning algorithms handling multivariate signals are able to capture the temporal as well as the spatial coherence of the EEG datasets \citep{Barthelemy2013a}. In this section, multivariate dictionaries are learned with M-DLA using the datasets of BCI Competition IV set 2a \citep{Brunner2008,TAN12}.
This is a motor imagery task, where 9 subjects have been instructed to imagine four classes of movements (left hand, right hand, tongue or feet) during two sessions conducted on different days.
Each session consists of 288 trials (72 trials for each class) and a trial is the 3 seconds recording of $\Vdim=22$ electrodes sampled at 250 Hz.
For each session, for each subject and for each task, the M-DLA learns a dictionary of five shiftable patterns (giving a five time redundant dictionary), with a sparsity parameter $K=1$. More details about this learning can be found in \citet{Barthelemy2013a}.

In the following, two experiments are presented which rely on the set-metrics between dictionaries to investigate the properties of the BCI dataset thanks to clustering techniques.
A part of the BCI community is organized and structured around the BCI Competition datasets, even if the individual variability of the subjects, both in term of inter-session evolution and in subject-specific characterization, is still largely unharnessed: between 20-30\% of the BCI subjects obtain catastrophic results with the state of the art algorithms, a phenomenon called ``BCI illiteracy'' or ``BCI-inefficiency'' \citep{VID10,HAM12}.
Is it possible to detect subjects who are ``BCI-inefficient''?
Do the subjects in the different competition datasets have the same variability? 
How important is the inter-session variability, knowing that a separate session is often used as an evaluation set for competition? 
All these questions require an objective measurement and set-metrics applied on learned dictionaries offer such a solution.
The present experiments are intended to demonstrate that the proposed set-metric are ready to be applied and that their immediate application offer some interesting insight on BCI dataset.

The first experiment proposes to investigate the link between the clustering of subjects and their BCI performance, in order to gain a better understanding of the ``BCI-inefficiency'' phenomenon. The second experiment is oriented to investigate the subject-specific intersession variability and its influence on the dictionary learning.

\subsection{Relation between Subject Clusters and BCI-Inefficiency}
\label{sec:illiteracy}

For these experiments, a dictionary is learned with M-DLA for each class of each subject: each subject is associated with four dictionaries (left hand, right hand, tongue and feet) learned over the first session of the dataset (used as a training set for the BCI Competition).

A distance matrix $G^c$ between subject is computed for each class $c$: for a given class all the distances between subject's dictionary are evaluated with the Wasserstein set-metric based on chordal distance.
%Using the Frobenius distance affect the results
All the distances $g^c_{ij}$ between subjects $i$ and $j$ are converted in Gaussian similarities $s_{ij}^c$ with the following relation : $s^c_{ij} = \exp\left(-(g^c_{ij})^2/2\sigma^2\right)$ with $\sigma=1$.
For each class, clusters of subjects are gathered using affinity propagation \citep{FRE07}.
Affinity propagation find the optimal number of clusters relying only on the similarities between subjects and the preference value of each subject, which the predisposition of each subject to become the exemplar of a cluster.
Here, we apply a common approach where all subjects have an identical preference value equal to the median of $G^c$.

\begin{figure}
  \centering
  % \resizebox{\linewidth}{!}{\input clusterSubjectScore-8-consensus5-dendrogram.tex}
  \pgfimage[interpolate=true,width=\linewidth]{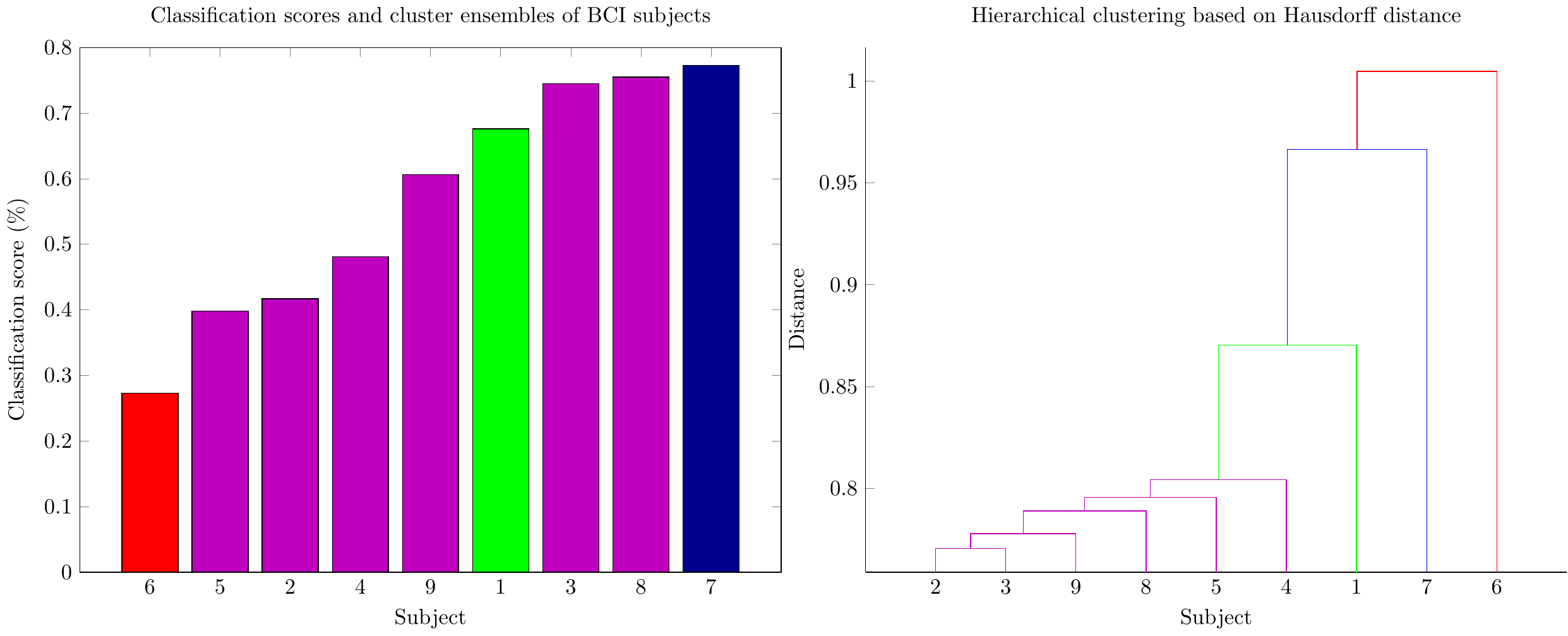}
  \caption{Left: Performance of the subjects from the BCI Competition IV-2a with state of the art algorithm from \citet{ANG12}, color indicates cluster ensemble obtained with consensus clustering on learned dictionaries.
Right: Hierarchical clustering on the same dataset, using Hausdorff distance to determine linkage.}
  \label{fig:bciressuj}
\end{figure}

The obtained cluster for each class are then combined using the cluster ensembles approach defined by \citet{STR03b}.
The results is a unique and global set of $C$ clusters, $C$ being a parameter, obtained by maximizing the shared mutual information of all classes.
A thorough analysis shows that subjects are aggregated in 3 or 4 stable cluster ensembles, represented with different colors on the left side of Figure~\ref{fig:bciressuj}.
On this Figure, the performance of each subject obtained with the state of the art algorithm \citep{ANG12} is shown on the ordinate-axis, the subjects are sorted according to their performance on the abscissa-axis.
Using $C=3$ cluster ensembles, it appears that the subject with the highest performance is always alone in a cluster, as the subject with the worst performance.
All the other subjects are gathered in the same cluster. 
With $C=4$, 5 or 6, only the four cluster ensembles shown on Figure~\ref{fig:bciressuj}(Left) are found.

To have a better visualization of the relation between cluster, a hierarchical clustering is shown on the right side of Figure~\ref{fig:bciressuj}.
It should be noticed that instead of applying a common linkage criteria, such as the single linkage or the complete linkage, we rely here on the Hausdorff distance to evaluate the distance between sets of subjects.
The results are very similar to those obtained with complete linkage.
On the resulting dendrogram, it is visible that the subjects \#2, \#3, \#9, \#8, \#5 and \#4 belong to the same cluster and that each of the subjects \#1, \#7 and \#6 constitute three separate clusters.

These experiments explicit new techniques to investigate the BCI inefficiency thanks to the set-metrics and the dictionary learning.
In the BCI Competition IV dataset 2a it appears that most of the subjects seem to share a common profile except two extreme cases, a BCI inefficient and a BCI efficient ones.
The possibilities offered by multivariate dictionary learning in conjunction with set-metric bring new opportunities to qualitatively assess  datasets used in competitions and challenges.
These approaches could help the community to propose more consistent and more complete benchmarks or evaluations.

\subsection{Temporal Variability of Subject Brain Waves}
\label{sec:tempvar}

In this section, the experiments aimed at investigating the temporal variability of a BCI subject, in other words how different are our brain waves from day to day?
Appreciating this variability is crucial as it is very common to train a classifier with session ``A'' and to evaluate the same classifier on data acquired from session ``B'', recorded several days after session ``A''.
This was the case for the BCI Competition IV dataset 2a.
Intersession variability is also at the heart of the calibration problem: a common practice is to calibrate the electronic BCI system, including the classifiers, each day.
Nonetheless a legitimate question when using a BCI system all day long is how often does it need to be calibrated?
The intrinsic temporal evolution of brain waves is a long studied problem and yet no reference methodology has been found.
The proposed experiments could help to properly design such methodology.

Here, a dictionary is learned with M-DLA for each subject, each class and each session.
Thus for each session, ``A'' and ``B'', there is 36 dictionaries learned, four by subject.
The parameters used for M-DLA are the same than the previous section.
The distances between dictionaries are computed with Wasserstein set-metric based on different ground distances: geodesic, chordal,  Fubini-Study and Binet-Cauchy, defined respectively in Equations~\eqref{eq:geodwong}, \eqref{eq:chordaldef}, \eqref{eq:fubinistudy} and~\eqref{eq:binetcauchy}.
As explained before, the clustering on dictionaries of each subject for a given class is obtained with the affinity propagation.
The clusters for each class are then combined in cluster ensembles with the consensus algorithm of \citet{STR03b}.
There are cluster ensembles for each ground distance and we also compute a global cluster ensembles gathering all classes and all ground distances for a given subject.

\begin{figure}
  \centering
  % \resizebox{0.75\linewidth}{!}{\input clusterSessionPlot.tex}
  \pgfimage[interpolate=true,width=0.75\linewidth]{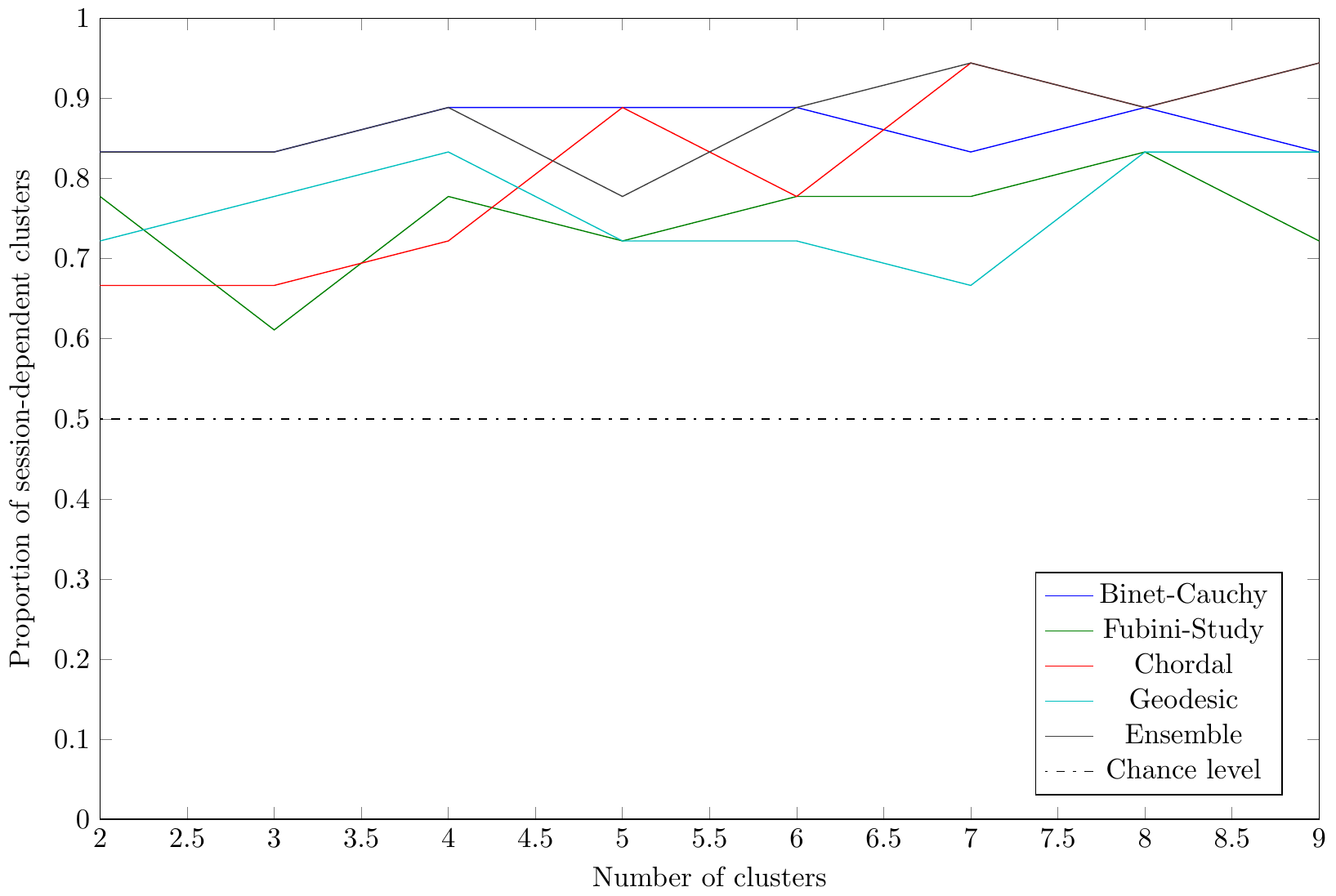}
  \caption{Consensus clustering on the learned dictionaries with different ground metrics. The proportion of clusters specific to one session is plotted as a function of the number of clusters.}
  \label{fig:clustsessionplot}
\end{figure}

The 9 subjects are separated based on the time of their recordings, either in session ``A'' and ``B'', resulting in 18 recordings to cluster.
A cluster could gather recordings from sessions ``A'' or ``B'' and is said to belong to the session which has the most important representation among the recordings.
For example, if a cluster gathers 3 recordings, two from session ``A'' and one from ``B'', the cluster is said to belong to ``A'' .
We propose to use as measure of dissimilarity the proportion of recordings in the same cluster.
A score of 1 indicates that all clusters contain either recordings from session ``A'' or, in the exclusive sense, from session ``B''.
The minimal score is 0 and indicates that each cluster contains as much recordings from session ``A'' and ``B''.
The chance level is at 0.5.

The Figure~\ref{fig:clustsessionplot} shows the proportion of cluster's recordings in the same session when the number $C$ of cluster ensembles increases. 
The different metrics yield the similar results and the global cluster ensembles, computed from all ground distances, displays a high level of dissimilarity between the two sessions.
Those results are stable with respect to the number of cluster ensembles $C$.

\begin{figure}[t!!]
  \centering
  \pgfimage[interpolate=true,width=0.9\linewidth]{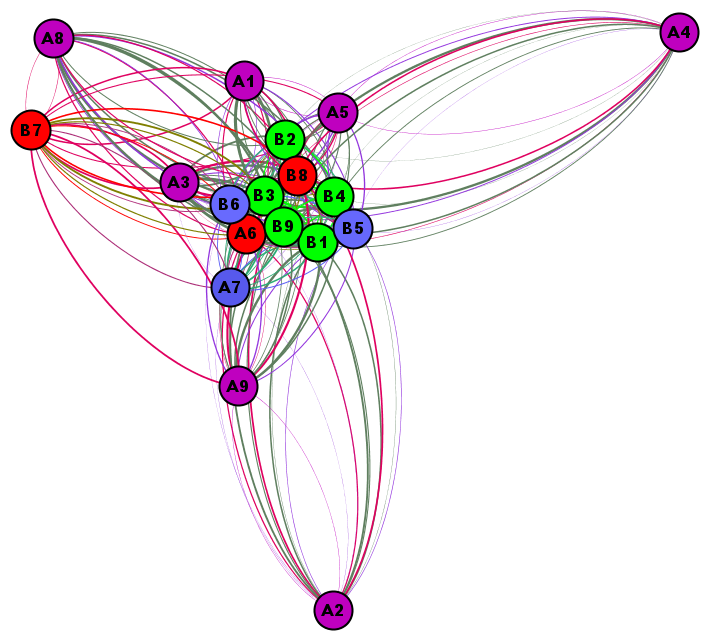}
  % \resizebox{\linewidth}{!}{\input clusterSessionLeigsv1cut.pdf_tex}
  % \pgfimage[interpolate=true,width=\linewidth]{Figures/sessionClusterAlternative}
  \caption{The 9 subjects recorded during 2 different sessions (A and B). The spatial positions chosen according to Laplacian eigenmaps. The color indicates the 4 cluster ensembles obtained with consensus clustering on learned dictionaries. The thickness of the links is proportional to the distance between learned dictionaries.}
  \label{fig:clustsession}
\end{figure}

The Figure~\ref{fig:clustsession} proposes a different visualization of the same phenomenon.
The 9 recordings of session ``A'' (one per subject) and the 9 ones of session ``B'' are represented with spatial positions drawn according to Laplacian eigenmaps \citep{BEL03}.
By applying an eigenvalue decomposition on the adjacency graph build from the distance matrix, the Laplacian eigenmaps embed the recording in $\Re^2$ and thus offer an isometric representation.
A thorough experimental analysis with the different metrics and the embedding produced by Laplacian eigenmaps leads us to choose $N=10$ as the number of nearest neighbors picked by the Laplacian eigenmaps and discrete weights ($t=\infty$).
Those parameters, used on the Fubini-Study distance matrix in the Figure~\ref{fig:clustsession}, have been chosen for their illustrative potential and different parameters yield qualitatively similar results.
The color code indicates how the recordings are separated into 4 cluster ensembles. 
The size of each link between recordings is proportional to the Wasserstein distance.

This figure shows that only two recordings (subjects \#6 and \#7 from session ``A'') are associated with cluster ensembles containing mainly recordings from session ``B''.
On the one hand, it appears clearly that all the recordings from session ``B'' are clustered and are separated by small distances, with the notable exception of the subject \#7.
On the other hand, recordings from session ``A'' are all very distant from recording of session ``B'' and belong all to the same cluster, except for subjects \#6 and \#7 which are part of clusters of session ``B''.
The subject \#7 is the one with the highest recognition rate, as shown on Figure~\ref{fig:bciressuj}: a tentative explanation is that the recordings of subject \#7 are very similar between sessions ``A'' and ``B'' whereas all others subjects recording differ from one session to the other.

As a corollary, when using a measure reflecting the subject-dependence of the results, no clear results appear and the measure stays at the chance level.
The measure used evaluates the dissimilarity between the subjects: each time the two sessions of a given subject are not in the same cluster, the indicator is decreased.

This demonstrates that there is a clear variation between sessions and that the current dictionary learning algorithm employed in these experiments, M-DLA, is sensitive to these session-dependent variations. The variations could be caused by the electronic system or could originate from the subject.
A first conclusion is that a BCI system should be recalibrated at the beginning of each session to avoid a performance drop.
The second conclusion concerns the dictionary learning algorithm: a training dataset including trials from several sessions could reinforce the robustness of the dictionary.

\begin{figure}
  \centering
  \pgfimage[interpolate=true,width=0.5\linewidth]{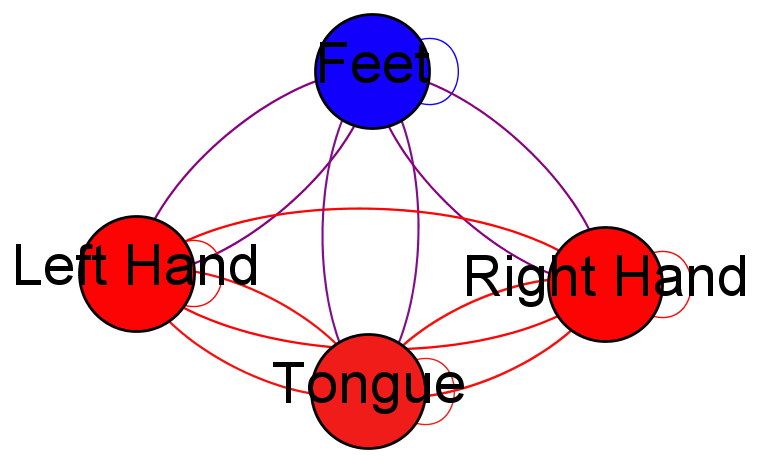}
  \caption{Consensus clustering on motor imagery classes: the cluster ensemble reflects the neurobiological organization of the brain.}
  \label{fig:bciresclass}
\end{figure}

The last experiment concerns the classes of the motor imagery task.
This time the distances are computed between classes for each subject, instead of computing distances between subjects for each class as in the previous experiment.
The M-DLA is applied with the same parameters, the affinity propagation is applied on the classes for each subject and each ground metric (Binet-Cauchy, Fubini-Study, chordal and geodesic).
Then the clusters of all subjects and metrics are merged in cluster ensembles with the consensus algorithm.
The resulting cluster ensembles are shown on Figure~\ref{fig:bciresclass}: the hands and tongue imagery are gathered in the same cluster.
The feet are set in a separate cluster.
This clustering reflect the neurobiological organization of the brain: on the one hand the localization of the cortical areas devoted to hand and tongue movements are situated on the sides of the head, in the parietal areas.
On the other hand, the surface of the motor cortex dedicated to the feet movements is situated on the interior side of the interhemispheric fissure, generating EEG components centered on the top of the head.

These experiments demonstrate that, properly equipped with a metric, the dictionary learning offer an efficient and robust approach to assess the intrinsic properties of multivariate datasets.
The formalization of a metric space applicable to frames open several direct applications, as shown in this section.

%%%%%%%%%%%%%%%%%%%%%%%%%%%%%%%%%%%%%%%%%%%%%%%%%%%%%%%%%%%%%%%%%%%

\section{Conclusion}
\label{sec:ccl}

% The overcomplete representations generated by dictionary learning offer a sound and powerful theoretical framework inherited from frame theory and matrix manifold.
% Nonetheless, the properties and structure of the dictionary space have not been fully explored, even if the interest for a formal definition of an associated metric space has kept growing.

This contribution rely on advances from algebraic geometry and frame theory to define suited metrics for multivariate dictionaries. 
It is the first contribution to define such metric spaces for learned dictionaries.
The distance between dictionaries is computed as the distance between their subspaces, yielding a pseudo-metric space which is invariant to linear transformations, an essential property to compare multivariate dictionaries.

The interest of the described set-metric is shown through its direct application on two examples: a synthetic dataset and real dataset of EEG for Brain Computer Interfaces. The defined metrics allow to estimate empirically the convergence of a dictionary learning algorithm with a precision outperforming the classical measurements based on detection rates. Furthermore, they also allow to qualitatively assess a specific dataset as it is shown with the BCI experiments.

In experiments on synthetic and real datasets, it is shown that the set-metrics could be adapted to the specific properties of the considered data, as adding invariance to rotations by choosing  a suited ground distance.  Indeed, the choice of the ground distance offers a large amount of possibilities to capture some specific properties linked to the properties of the underlying data. Ground metrics such as the chordal or the geodesic distance are a good default choice. %Prospects are to investigate ground distances invariant only to sign or only to rotation.

The chosen examples amount from multivariate data analysis, illustrating that the proposed metrics could be applied straightly on these complex problems.
Nonetheless, all the work described in this contribution is also valid for the univariate case and operates with every dictionary learning algorithms.
But in numerous applications, multivariate datasets are transformed to be processed by univariate algorithms, causing an important information loss.
Metrics for multivariate dictionaries bring a framework to develop new and improved multivariate dictionary learning algorithms.

At least, an original connection is made between Grassmannian frames, compressed sensing and dictionary learning.
This disjoint research topics try either to cover the entire space under consideration, as in packing problems, or to minimize the reconstruction error on the signal energy, as in dictionary learning, or to add coherence constraints on dictionary learning algorithms, as in hybrid approaches.

\acks{Sylvain Chevallier was partially supported by a grant from EADS Foundation, project Cerebraptic.}

%%%%%%%%%%%%%%%%%%%%%%%%%%%%%%%%%%%%%%%%%%%%%%%%%%%%%%%%%%%%%%%%%%%
\newpage
\appendix
\section*{Appendix A.}
\label{app:manifold}

In this appendix, we explicit the definitions of manifolds, matrix manifolds and Grassmannian Manifold.

\subsection*{Manifolds}
% \paragraph{Manifolds}

\begin{definition}
Let us consider an $\N$-dimensional vector space $\vecspace$ with inner product $\langle \cdot, \cdot \rangle$ defined over the field $\field$. A \emph{manifold} is a topological space $\mc{M}$ of class $C^k$ (differentiable up to $k^{\tmop{th}}$ degree of differentiation) together with an \emph{atlas} $\mc{A} = (\phi_i, Z_i)_{i\in I}$ on $\vecspace$, with $I \subset \mathbb{N}$ an indexing set.
An atlas is a collection of charts, a chart $(\phi_i, Z_i)$ being an homeomorphism of the open subset $Z_i \subset \mc{M}$ onto an open subset $\phi(Z_i) \subset \vecspace$, that is $(\phi_i, Z_i)$ is a bijection, is continuous and its inverse $\phi^{-1}$ is continuous.
\end{definition}

A manifold is differentiable if it endows a globally defined differentiable structure.
To induce a globally differentiable structure from the charts, the composition of intersecting charts must be a differentiable function and the charts are said $C^k$-compatible.
For example if two charts overlap, the coordinates defined by a chart are required to be differentiable with respect to the coordinates defined in the other chart.
Formally, any two charts $(\phi_i, Z_i)$, $(\phi_j,Z_j)$ are $C^k$-compatible if
\begin{equation*}
  \label{eq:chartoverlap}
  \phi_{ij} := \{ \phi_i \circ \phi_j^{-1}|_{\phi(Z_i\cap Z_j)} : \phi_j (Z_i \cap Z_j) \ra \phi_i(Z_i\cap Z_j)  \} \ ,
\end{equation*}
and its inverse $\phi_{ji}$ are of class $C^k$.
If $\mc{M}$ is $C^k$ for all $k$, $\mc{M}$ is said smooth or $C^\infty$-manifold.

\subsection*{Matrix Manifolds}
% \paragraph{Matrix manifolds}
\label{sec:matmani}

\begin{figure}
  \centering
  \resizebox{0.5\linewidth}{!}{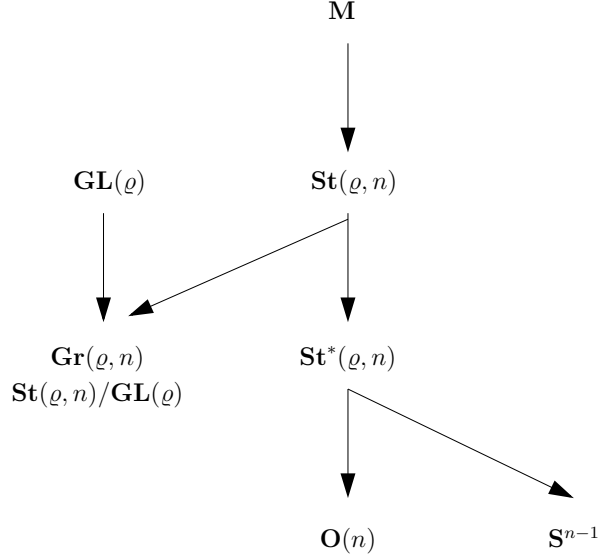}
  \caption{Hierarchy of matrix manifolds}
  \label{fig:hierarchy}
\end{figure}

First, remark that if we let $\{e_i\}_{i=1}^\N$  be a basis of $\vecspace$, then the function 
\begin{equation*}
  \label{eq:chartlinman}
 \phi: \vecspace \ra \field^\N , \fone \mapsto [\fone_1, \ldots, \fone_\N]
\end{equation*}
such that $\fone = \sum_{i=1}^\N \fone_ie_i$ is a chart of the set $\vecspace$.
As all the charts constructed in this way  are compatible (or ``smooth''), they form an atlas of the set $\vecspace$, endowing $\vecspace$ with a manifold structure.
Thus, every vector space is a linear manifold.

%For this section, we will consider only real matrices and their associated vector spaces.
The set $\field^{\N \times \V}$ with the usual sum and multiplication by a scalar is a vector space.
Hence, it is endowed with a linear manifold structure and a chart of this manifold is 
\begin{equation*}
  \label{eq:chartmatman}
  \phi : \field^{\N \times \V} \ra \field^{\N\V} \,, \Fone \mapsto \tmop{vec}(\Fone) \ ,
\end{equation*}
where $\tmop{vec}(\Fone)$ denotes the vectorization of the matrix $\Fone$.
The set $\field^{\N \times \V}$ together with its linear manifold structure is denoted $\mathbf{M}$ hereafter.
The manifold $\mathbf{M}$ is the set of all matrices in $\field^{\N \times \V}$, that is the set of all linear applications $\mc{L} : \field^\N \ra \field^\V$ or equivalently the set $\field^\V \otimes \field^\N$ defined as the tensor product of $\field^\V$ and $\field^\N$.
Also, $\mathbf{M}$ can be seen as an Euclidean space with the inner product $\langle \Fone, \Ftwo \rangle := \tmop{vec}(\Fone)^T\tmop{vec}(\Ftwo) = \tmop{trace}(\Fone^T\Ftwo)$ and the induced norm $\norm{\Fone}_F$.

\begin{definition}
  Let $\N$ and $\V$ a positive integer with $\V \leqslant \N$.
  The \emph{nonorthogonal Stiefel manifold $\St(\V,\N)$} is the set of all full rank matrices in $\mathbf{M}$, that is
  \begin{displaymath}
    \St(\V,\N) := \{ \Fone \in \mathbf{M} : \tmop{rank}(\Fone) = \V \} \ .
  \end{displaymath}
\end{definition}

The column space $\subspaceFone$ of a matrix $\Fone$ in $\St(\varrho,\N)$ defines the basis of a $\varrho$-dimensional subspace in $\field^\N$, that is the $\varrho$ column vectors ($\fone_1, \ldots, \fone_\varrho$) of $\Fone$ span the subspace $\subspaceFone$, thus $\tmop{span}(\Fone)=\subspaceFone$.
For $\N=\varrho$, the Stiefel manifold $\St(\varrho,\varrho)$ reduces to the general linear group $\GL(\varrho)$.

Stiefel manifold defined here is different from the orthogonal Stiefel manifold $\St^*$, which is the set of orthonormal $\N$-by-$\V$ matrices.
\begin{definition} 
Let $\N$ and $\V$ be two positive integers such that $\V \leqslant \N$.
The \emph{orthogonal Stiefel manifold} $\St^*(\V,\N)$ is defined as
\begin{displaymath}
  \St^*(\V,\N) := \{ \Fone \in \mathbf{M} : \Fone^T\Fone = I_\V \} \ ,
\end{displaymath}with $I_\V$ the $\V$-by-$\V$ identity matrix. 
\end{definition}
For $\V=1$, $\St^*(1,\N)$ reduces to the unit sphere $\mathbf{S}^{\N-1}$ and for $\V=\N$, it reduces to the set of orthogonal matrices $\Or(\N)$, as shown on Figure~\ref{fig:hierarchy}.

\subsection*{Grassmannian Manifold}

\begin{figure}
  \centering
  \resizebox{0.3\linewidth}{!}{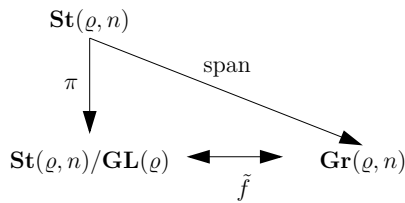}
  \caption{Construction of the Grassmannian manifold. Here, the projection $\pi$ verifies $\pi^{-1}(\pi(\Fone))=\{\Fone\GL(\V) : \Fone \in \GL(\V)\}$ and $\tilde{\fone}$ is the projection of $\fone$ on $\Gr(\V,\N)$ such that $\fone=\tilde{\fone}\circ\pi$.}
  \label{fig:grass}
\end{figure}

The Grassmannian manifold $\Gr(\V,\N)$ is a quotient manifold, such as an element $\subspaceFone$ of $\Gr(\V,\N)$ can also be caracterized as a $\V$-dimensional subspace in $\field^\N$.
A matrix $\Fone$ is said to span $\subspaceFone$ if $\tmop{span}(\Fone)=\subspaceFone$, and the span of $\Fone$ is an element of $\Gr(\V,\N)$ only if $\Fone \in \St(\V,\N)$.
There is an infinite number of matrices in $\St(\V,\N)$ which span an element $\subspaceFone$ of $\Gr(\V,\N)$, as illustrated on Figure~\ref{fig:grass}.

\begin{definition}
Given a matrix $\Fone \in \St(\V,\N)$, the set of the matrices that have the same span as $\Fone$ is defined as
\begin{equation*}
  \label{eq:spaneq}
  \Fone\GL(\V)  := \{ \Fone A : A \in \GL(\V)\} \ ,
\end{equation*}
where $\GL(\V)$ is the set of $\V$-by-$\V$ invertible matrices.
The \emph{Grassmannian manifold} $\Gr(\V,\N)$ is defined as the orbit space of 
\begin{equation*}
  \label{eq:grassmann}
  \St(\V,\N)/\GL(\V) := \{ \Fone\GL(\V) : \Fone \in \St(\V,\N)\} \ ,
\end{equation*}
with the group action $\St(\V,\N)\times \GL(\V) \ra \St(\V,\N), (\Fone, A) \mapsto \Fone A$.
\end{definition}

In the case $\N=2$ and $\V=1$, an element of a Grassmannian manifold defines an equivalence line in the plane, that is a set of points in $\St(\V,\N)$ which are invariant under the group $\GL(\V)$.
In a more general case, each element $\subspaceFone$ of a $\Gr(\V,\N)$ defines a subspace and for each $\subspaceFone$ corresponds an equivalent class of matrices that span $\subspaceFone$.
 
% Stiefel and Grassmannian manifolds mainly differ by the order and the choice of the basis.
% As Grassmannian are quotient manifold, thus defined from an equivalence relation, there is no unique order or basis of a matrix representing an element of the manifold as each element describes a class of equivalent matrices.
% Hence, by definition Grassmannian are invariant to permutations.
% Conversely, the order of the basis is important for the Stiefel manifolds.

%%%%%%%%%%%%%%%%%%%%%%%%%%%%%%%%%%%%%%%%%%%%%%%%%%%%%%%%%%%%%%%%%%%

\vskip 0.2in
\bibliographystyle{plainnat}
\bibliography{frames}

\begin{thebibliography}{78}
\providecommand{\natexlab}[1]{#1}
\providecommand{\url}[1]{\texttt{#1}}
\expandafter\ifx\csname urlstyle\endcsname\relax
  \providecommand{\doi}[1]{doi: #1}\else
  \providecommand{\doi}{doi: \begingroup \urlstyle{rm}\Url}\fi

\bibitem[Aharon(2006)]{Aharon2006b}
M.~Aharon.
\newblock \emph{Overcomplete Dictionaries for Sparse Representation of
  Signals}.
\newblock PhD thesis, Technion - Israel Institute of Technology, 2006.

\bibitem[Aharon et~al.(2006)Aharon, Elad, and Bruckstein]{Aharon2006}
M.~Aharon, M.~Elad, and A.M. Bruckstein.
\newblock {K-SVD}: An algorithm for designing overcomplete dictionaries for
  sparse representation.
\newblock \emph{IEEE Trans. Signal Processing}, 54:\penalty0 4311--4322, 2006.

\bibitem[Aldroubi(1995)]{ALD95}
A.~Aldroubi.
\newblock {Portraits of Frames}.
\newblock \emph{Proceedings of the American Mathematical Society}, 123\penalty0
  (6):\penalty0 1661--1668, 1995.

\bibitem[Ang et~al.(2012)Ang, Chin, Wang, Guan, and Zhang]{ANG12}
K.K. Ang, Z.Y. Chin, C.~Wang, C.~Guan, and H.~Zhang.
\newblock Filter bank common spatial pattern algorithm on {BCI} competition
  {IV} datasets 2a and 2b.
\newblock \emph{Frontiers in Neuroscience}, 6:\penalty0 1--9, 2012.

\bibitem[Applebaum et~al.(2009)Applebaum, Howard, Searle, and
  Calderbank]{APP09}
L.~Applebaum, S.D. Howard, S.~Searle, and R.~Calderbank.
\newblock Chirp sensing codes: Deterministic compressed sensing measurements
  for fast recovery.
\newblock \emph{Applied and Computational Harmonic Analysis}, 26\penalty0
  (2):\penalty0 283--290, 2009.

\bibitem[Balan(1999)]{BAL99}
R.~Balan.
\newblock {Equivalence Relations and Distances between Hilbert Frames}.
\newblock \emph{Proceedings of the American Mathematical Society}, 127\penalty0
  (8):\penalty0 2353--2366, 1999.

\bibitem[Barth\'elemy et~al.(2012)Barth\'elemy, Larue, Mayoue, Mercier, and
  Mars]{Barthelemy2012}
Q.~Barth\'elemy, A.~Larue, A.~Mayoue, D.~Mercier, and J.I. Mars.
\newblock Shift \& {2D} rotation invariant sparse coding for multivariate
  signals.
\newblock \emph{IEEE Trans. Signal Processing}, 60:\penalty0 1597--1611, 2012.

\bibitem[Barth\'elemy et~al.(2013{\natexlab{a}})Barth\'elemy, Gouy-Pailler,
  Isaac, Souloumiac, Larue, and Mars]{Barthelemy2013a}
Q.~Barth\'elemy, C.~Gouy-Pailler, Y.~Isaac, A.~Souloumiac, A.~Larue, and J.I.
  Mars.
\newblock Multivariate temporal dictionary learning for {EEG}.
\newblock \emph{Journal of Neuroscience Methods}, in press, 2013{\natexlab{a}}.

\bibitem[Barth\'elemy et~al.(2013{\natexlab{b}})Barth\'elemy, Larue, and
  Mars]{Barthelemy2013}
Q.~Barth\'elemy, A.~Larue, and J.I. Mars.
\newblock Decomposition and dictionary learning for {3D} trajectories.
\newblock Technical report, CEA, 2013{\natexlab{b}}.

\bibitem[Beck and Teboulle(2009)]{BEC09}
A.~Beck and M.~Teboulle.
\newblock {A Fast Iterative {Shrinkage-Thresholding} Algorithm for Linear
  Inverse Problems}.
\newblock \emph{SIAM J. Img. Sci.}, 2\penalty0 (1):\penalty0 183--202, 2009.

\bibitem[Belkin and Niyogi(2003)]{BEL03}
M.~Belkin and P.~Niyogi.
\newblock Laplacian eigenmaps for dimensionality reduction and data
  representation.
\newblock \emph{Neural Comput.}, 15\penalty0 (6):\penalty0 1373--1396, 2003.

\bibitem[Brunner et~al.(2008)Brunner, Leeb, M\"uller-Putz, Schl\"ogl, and
  Pfurtscheller]{Brunner2008}
C.~Brunner, R.~Leeb, G.R. M\"uller-Putz, A.~Schl\"ogl, and G.~Pfurtscheller.
\newblock {BCI Competition 2008 - Graz data set A}, 2008.

\bibitem[Burago et~al.(2001)Burago, Burago, and Ivanov]{buragobook}
D.~Burago, Y.~Burago, and S.~Ivanov.
\newblock \emph{A course in metric geometry}.
\newblock American Mathematical Society Providence, 2001.

\bibitem[Cand{\`e}s and Donoho(2004)]{CAN04}
E.J. Cand{\`e}s and D.L. Donoho.
\newblock {New tight frames of curvelets and optimal representations of objects
  with piecewise C2 singularities}.
\newblock \emph{Comm. Pure Appl. Math.}, 57\penalty0 (2):\penalty0 219--266,
  2004.

\bibitem[Cand\`{e}s et~al.(2006)Cand\`{e}s, Romberg, and Tao]{CAN06}
E.J. Cand\`{e}s, J.K. Romberg, and T.~Tao.
\newblock Stable signal recovery from incomplete and inaccurate measurements.
\newblock \emph{Comm. Pure Appl. Math.}, 59\penalty0 (8):\penalty0 1207--1223,
  2006.

\bibitem[Cand{\`e}s et~al.(2006)Cand{\`e}s, Romberg, and Tao]{CAN06a}
E.J. Cand{\`e}s, J.K. Romberg, and T.~Tao.
\newblock Robust uncertainty principles: exact signal reconstruction from
  highly incomplete frequency information.
\newblock \emph{IEEE Trans. Information Theory}, 52\penalty0 (2):\penalty0
  489--509, 2006.

\bibitem[Casazza and Kutyniok(2004)]{CAS04}
P.G. Casazza and G.~Kutyniok.
\newblock \emph{Frames of subspaces}, pages 87--114.
\newblock Wavelets, Frames and Operator Theory. AMS, 2004.

\bibitem[Chikuse(2003)]{CHI03}
Y.~Chikuse.
\newblock \emph{{Statistics on Special Manifolds}}.
\newblock Springer, 2003.

\bibitem[Christensen(2003)]{christensen2003introduction}
O.~Christensen.
\newblock \emph{{An introduction to frames and Riesz bases}}.
\newblock Birkhauser, 2003.

\bibitem[Conway et~al.(1996)Conway, Hardin, and Sloane]{CON96}
J.H. Conway, R.H. Hardin, and N.J.A. Sloane.
\newblock Packing lines, planes, etc.: Packings in grassmannian spaces.
\newblock \emph{Experimental Mathematics}, 5\penalty0 (2):\penalty0 139--159,
  January 1996.

\bibitem[Daubechies et~al.(1986)Daubechies, Grossmann, and Meyer]{DAU86}
I.~Daubechies, A.~Grossmann, and Y.~Meyer.
\newblock Painless nonorthogonal expansions.
\newblock \emph{Journal of Mathematical Physics}, 27\penalty0 (5):\penalty0
  1271--1283, 1986.

\bibitem[Davis(1994)]{Davis1994a}
G.~Davis.
\newblock \emph{Adaptive Nonlinear Approximations}.
\newblock PhD thesis, New York University, 1994.

\bibitem[DeVore(2007)]{DEV07}
R.A. DeVore.
\newblock Deterministic constructions of compressed sensing matrices.
\newblock \emph{Journal of Complexity}, 23\penalty0 (4-6):\penalty0 918--925,
  2007.

\bibitem[Dhillon et~al.(2008)Dhillon, Heath~Jr., Strohmer, and Tropp]{DHI08}
I.S. Dhillon, R.W. Heath~Jr., T.~Strohmer, and J.A. Tropp.
\newblock {Constructing Packings in Grassmannian Manifolds via Alternating
  Projection}.
\newblock \emph{Experimental Mathematics}, 17\penalty0 (1):\penalty0 9--35,
  2008.

\bibitem[Donoho(2006)]{Donoho2006}
D.L. Donoho.
\newblock Compressed sensing.
\newblock \emph{IEEE Trans. Information Theory}, 52:\penalty0 1289--1306, 2006.

\bibitem[Duffin and Schaeffer(1952)]{DUF52}
R.J. Duffin and A.C. Schaeffer.
\newblock A class of nonharmonic {Fourier} series.
\newblock \emph{Trans. Amer. Math. Soc.}, 72\penalty0 (2):\penalty0 341--366,
  1952.

\bibitem[Edelman et~al.(1999)Edelman, Arias, and Smith]{EDE99}
A.~Edelman, T.A. Arias, and S.T. Smith.
\newblock {The Geometry of Algorithms with Orthogonality Constraints}.
\newblock \emph{SIAM J. Matrix Anal. Appl.}, 20\penalty0 (2):\penalty0
  303--353, 1999.

\bibitem[Efron et~al.(2004)Efron, Hastie, Johnstone, and Tibshirani]{EFR04}
B.~Efron, T.~Hastie, L.~Johnstone, and R.~Tibshirani.
\newblock {Least Angle Regression}.
\newblock \emph{The Annals of Statistics}, 32\penalty0 (2):\penalty0 407--451,
  2004.

\bibitem[Engan et~al.(2000)Engan, Aase, and Hus{\o}y]{Engan2000}
K.~Engan, S.O. Aase, and J.H. Hus{\o}y.
\newblock Multi-frame compression: theory and design.
\newblock \emph{Signal Process.}, 80:\penalty0 2121--2140, 2000.

\bibitem[Engan et~al.(2007)Engan, Skretting, and Hus{\o}y]{Engan2007}
K.~Engan, K.~Skretting, and J.H. Hus{\o}y.
\newblock Family of iterative {LS}-based dictionary learning algorithms,
  {ILS-DLA}, for sparse signal representation.
\newblock \emph{Digit. Signal Process.}, 17:\penalty0 32--49, 2007.

\bibitem[Frey and Dueck(2007)]{FRE07}
B.J. Frey and D.~Dueck.
\newblock Clustering by passing messages between data points.
\newblock \emph{Science}, 315:\penalty0 972--976, 2007.

\bibitem[Golub and van Loan(1996)]{GOL96}
G.H. Golub and C.F. van Loan.
\newblock \emph{Matrix Computations}.
\newblock The Johns Hopkins University Press, 3rd edition, 1996.

\bibitem[{Golub} and {Zha}(1995)]{GOL95}
G.H. {Golub} and H.~{Zha}.
\newblock The canonical correlations of matrix pairs and their numerical
  computation.
\newblock \emph{Institute for Mathematics and Its Applications}, 69:\penalty0
  27--49, 1995.

\bibitem[Gribonval and Schnass(2010)]{GRI10}
R.~Gribonval and K.~Schnass.
\newblock Dictionary {Identification--Sparse} {Matrix-Factorization} via
  {l1-Minimization}.
\newblock \emph{IEEE Trans. Information Theory}, 56\penalty0 (7):\penalty0
  3523--3539, 2010.

\bibitem[Gribonval et~al.(2007)Gribonval, Rauhut, Schnass, and
  Vandergheynst]{Gribonval2007}
R.~Gribonval, H.~Rauhut, K.~Schnass, and P.~Vandergheynst.
\newblock Atoms of all channels, unite! average case analysis of multi-channel
  sparse recovery using greedy algorithms.
\newblock Technical Report PI-1848, IRISA, 2007.

\bibitem[Grosse et~al.(2007)Grosse, Raina, Kwong, and Ng]{GRO07}
R.~Grosse, R.~Raina, H.~Kwong, and A.Y. Ng.
\newblock {Shift-Invariant Sparse Coding for Audio Classification}.
\newblock In \emph{{Proc. Conf. on Uncertainty in Artificial Intelligence
  UAI}}, 2007.

\bibitem[Hamm and Lee(2008)]{HAM08}
J.~Hamm and D.D. Lee.
\newblock {Grassmann discriminant analysis: a unifying view on subspace-based
  learning}.
\newblock In \emph{{Proc. of ICML}}, pages 376--383. ACM, 2008.

\bibitem[Hammer et~al.(2012)Hammer, Halder, Blankertz, Sannelli, Dickhaus,
  Kleih, M\"{u}ller, and K\"{u}bler]{HAM12}
E.M. Hammer, S.~Halder, B.~Blankertz, C.~Sannelli, T.~Dickhaus, S.~Kleih, K.-R.
  M\"{u}ller, and A.~K\"{u}bler.
\newblock Psychological predictors of {SMR}-{BCI} performance.
\newblock \emph{Biological Psychology}, 89\penalty0 (1):\penalty0 80--86, 2012.

\bibitem[Hotelling(1936)]{HOT36}
H.~Hotelling.
\newblock Relations between two sets of variates.
\newblock \emph{Biometrika}, 28\penalty0 (3/4):\penalty0 321--377, 1936.

\bibitem[Jenatton et~al.(2012)Jenatton, Gribonval, and Bach]{JEN12}
R.~Jenatton, R.~Gribonval, and F.~Bach.
\newblock Local stability and robustness of sparse dictionary learning in the
  presence of noise, 2012.

\bibitem[Kobayashi and Nomizu(1969)]{KOB69}
S.~Kobayashi and K.~Nomizu.
\newblock \emph{{Foundations of Differential Geometry}}.
\newblock Wiley-Interscience, 1969.

\bibitem[Kutyniok et~al.(2009)Kutyniok, Pezeshki, Calderbank, and Liu]{KUT09}
G.~Kutyniok, A.~Pezeshki, R.~Calderbank, and T.~Liu.
\newblock Robust dimension reduction, fusion frames, and grassmannian packings.
\newblock \emph{Applied and Computational Harmonic Analysis}, 26\penalty0
  (1):\penalty0 64--76, 2009.

\bibitem[Lesage(2007)]{LES07}
S.~Lesage.
\newblock \emph{{Apprentissage de dictionnaires structur{\'e}s pour la
  mod{\'e}lisation parcimonieuse des signaux multicanaux}}.
\newblock Th{\`e}se de {D}octorat, Universit{\'e} de Rennes, 2007.

\bibitem[Lui(2012)]{LUI12}
Y.M. Lui.
\newblock {Advances in matrix manifolds for computer vision}.
\newblock \emph{Image and Vision Computing}, 30\penalty0 (6-7):\penalty0
  380--388, 2012.

\bibitem[Mailh\'e et~al.(2009)Mailh\'e, Gribonval, Bimbot, Lemay,
  Vandergheynst, and Vesin]{Mailhe2009a}
B.~Mailh\'e, R.~Gribonval, F.~Bimbot, M.~Lemay, P.~Vandergheynst, and J.M.
  Vesin.
\newblock Dictionary learning for the sparse modelling of atrial fibrillation
  in {ECG} signals.
\newblock In \emph{IEEE ICASSP}, pages 465--468, 2009.

\bibitem[Mairal et~al.(2008)Mairal, Elad, and Sapiro]{Mairal2008}
J.~Mairal, M.~Elad, and G.~Sapiro.
\newblock Sparse representation for color image restoration.
\newblock \emph{IEEE Trans. Image Processing}, 17:\penalty0 53--69, 2008.

\bibitem[Mairal et~al.(2009)Mairal, Bach, Ponce, Sapiro, and Zisserman]{MAI09a}
J.~Mairal, F.~Bach, J.~Ponce, G.~Sapiro, and A.~Zisserman.
\newblock {Non-local sparse models for image restoration}.
\newblock In \emph{{IEEE ICCV}}, pages 2272--2279, 2009.

\bibitem[Mairal et~al.(2010)Mairal, Bach, Ponce, and Sapiro]{Mairal2010}
J.~Mairal, F.~Bach, J.~Ponce, and G.~Sapiro.
\newblock Online learning for matrix factorization and sparse coding.
\newblock \emph{J. Mach. Learn. Res.}, 11:\penalty0 19--60, 2010.

\bibitem[Mallat(1999)]{MAL99}
S.~Mallat.
\newblock \emph{{A Wavelet Tour of Signal Processing, Second Edition (Wavelet
  Analysis \& Its Applications)}}.
\newblock Academic Press, 1999.

\bibitem[Mallat and Zhang(1993)]{Mallat1993a}
S.G. Mallat and Z.~Zhang.
\newblock Matching pursuits with time-frequency dictionaries.
\newblock \emph{IEEE Trans. Signal Processing}, 41:\penalty0 3397--3415, 1993.

\bibitem[Meyer(1995)]{MEY95}
Y.~Meyer.
\newblock \emph{Wavelets and Operators: Volume 1}.
\newblock Cambridge University Press, 1995.

\bibitem[Milnor and Stasheff(1974)]{milnor-book}
J.W. Milnor and J.D. Stasheff.
\newblock \emph{Characteristic Classes.(AM-76)}, volume~76.
\newblock Princeton University Press, 1974.

\bibitem[Mixon et~al.(2011)Mixon, Quinn, Kiyavash, and Fickus]{MIX11b}
D.G. Mixon, C.J. Quinn, N.~Kiyavash, and M.~Fickus.
\newblock Equiangular tight frame fingerprinting codes.
\newblock In \emph{IEEE ICASSP}, pages 1856--1859, 2011.

\bibitem[Monaci et~al.(2007)Monaci, Jost, Vandergheynst, Mailh\'{e}, Lesage,
  and Gribonval]{MON07}
G.~Monaci, P.~Jost, P.~Vandergheynst, B.~Mailh\'{e}, S.~Lesage, and
  R.~Gribonval.
\newblock {Learning multimodal dictionaries}.
\newblock \emph{IEEE Trans. Image Processing}, 16:\penalty0 2272--2283, 2007.

\bibitem[Monaci et~al.(2009)Monaci, Vandergheynst, and Sommer]{Monaci2009}
G.~Monaci, P.~Vandergheynst, and F.T. Sommer.
\newblock Learning bimodal structure in audio-visual data.
\newblock \emph{IEEE Trans. Neural Networks}, 20:\penalty0 1898--1910, 2009.

\bibitem[Moudden et~al.(2009)Moudden, Bobin, Starck, and Fadili]{MOU09}
Y.~Moudden, J.~Bobin, J.-L. Starck, and J.~Fadili.
\newblock {Dictionary learning with spatio-spectral sparsity constraints}.
\newblock In \emph{SPARS '09}, 2009.

\bibitem[Pati et~al.(1993)Pati, Rezaiifar, and Krishnaprasad]{Pati1993}
Y.C. Pati, R.~Rezaiifar, and P.S. Krishnaprasad.
\newblock {Orthogonal Matching Pursuit}: recursive function approximation with
  applications to wavelet decomposition.
\newblock In \emph{Proc. Asilomar Conf. on Signals, Systems and Comput.}, 1993.

\bibitem[Rakotomamonjy(2011)]{Rakotomamonjy2011}
A.~Rakotomamonjy.
\newblock Surveying and comparing simultaneous sparse approximation (or
  group-lasso) algorithms.
\newblock \emph{Signal Process.}, 91:\penalty0 1505--1526, 2011.

\bibitem[Skretting and Engan(2010)]{SKR10}
K.~Skretting and K.~Engan.
\newblock Recursive least squares dictionary learning algorithm.
\newblock \emph{IEEE Trans. Signal Processing}, 58:\penalty0 2121--2130, 2010.

\bibitem[{Skretting} and {Engan}(2011)]{SKR11}
K.~{Skretting} and K.~{Engan}.
\newblock Learned dictionaries for sparse image representation: properties and
  results.
\newblock In \emph{SPIE Conference}, volume 8138, 2011.

\bibitem[Strawn(2012)]{STR12}
N.~Strawn.
\newblock Optimization over finite frame varietes and structured dictionary
  design.
\newblock \emph{Applied and Computational Harmonic Analysis}, 32:\penalty0
  413--434, 2012.

\bibitem[Strehl and Ghosh(2003)]{STR03b}
A.~Strehl and J.~Ghosh.
\newblock Cluster ensembles --- a knowledge reuse framework for combining
  multiple partitions.
\newblock \emph{J. Mach. Learn. Res.}, 3:\penalty0 583--617, 2003.

\bibitem[Strohmer and Heath~Jr.(2003)]{STR03}
T.~Strohmer and R.W. Heath~Jr.
\newblock Grassmannian frames with applications to coding and communication.
\newblock \emph{Applied and Computational Harmonic Analysis}, 14\penalty0
  (3):\penalty0 257--275, 2003.

\bibitem[Tammes(1930)]{TAM30}
P.~Tammes.
\newblock \emph{On the origin of number and arrangement of the places of exit
  on the surface of pollen-grains}.
\newblock PhD thesis, Faculty of Groningen, 1930.

\bibitem[Tangermann et~al.(2012)Tangermann, M\"uller, Aertsen, Birbaumer,
  Braun, Brunner, Leeb, Mehring, Miller, M\"uller-Putz, Nolte, Pfurtscheller,
  Preissl, Schalk, Schl\"ogl, Vidaurre, Waldert, and Blankertz]{TAN12}
M.~Tangermann, K.-R. M\"uller, A.~Aertsen, N.~Birbaumer, C.~Braun, C.~Brunner,
  R.~Leeb, C.~Mehring, K.J. Miller, G.~M\"uller-Putz, G.~Nolte,
  G.~Pfurtscheller, H.~Preissl, G.~Schalk, A.~Schl\"ogl, C.~Vidaurre,
  S.~Waldert, and B.~Blankertz.
\newblock Review of the {BCI Competition IV}.
\newblock \emph{Frontiers in Neuroscience}, 6\penalty0 (55), 2012.

\bibitem[Tibshirani(1996)]{TIB96}
R.~Tibshirani.
\newblock {Regression Shrinkage and Selection via the Lasso}.
\newblock \emph{Journal of the Royal Statistical Society. Series B
  (Methodological)}, 58\penalty0 (1):\penalty0 267--288, 1996.

\bibitem[To\v{s}i\'c and Frossard(2011{\natexlab{a}})]{Tosic2011}
I.~To\v{s}i\'c and P.~Frossard.
\newblock Dictionary learning.
\newblock \emph{IEEE Signal Processing Magazine}, 28:\penalty0 27--38,
  2011{\natexlab{a}}.

\bibitem[To\v{s}i\'c and Frossard(2011{\natexlab{b}})]{Tosic2011a}
I.~To\v{s}i\'c and P.~Frossard.
\newblock Dictionary learning for stereo image representation.
\newblock \emph{IEEE Trans. Image Processing}, 20:\penalty0 921--934,
  2011{\natexlab{b}}.

\bibitem[Tropp(2004)]{TRO04}
J.A. Tropp.
\newblock Greed is good: algorithmic results for sparse approximation.
\newblock \emph{IEEE Trans. Information Theory}, 50\penalty0 (10):\penalty0
  2231--2242, 2004.

\bibitem[Tropp and Wright(2010)]{Tropp2010}
J.A. Tropp and S.J. Wright.
\newblock Computational methods for sparse solution of linear inverse problems.
\newblock \emph{Proceedings of the IEEE}, 98:\penalty0 948--958, 2010.

\bibitem[Vainsencher et~al.(2011)Vainsencher, Mannor, and
  Bruckstein]{Vainsencher2011}
D.~Vainsencher, S.~Mannor, and A.M. Bruckstein.
\newblock The sample complexity of dictionary learning.
\newblock \emph{J. Mach. Learn. Res.}, 12:\penalty0 3259--3281, 2011.

\bibitem[Vidaurre and Blankertz(2010)]{VID10}
C.~Vidaurre and B.~Blankertz.
\newblock Towards a cure for {BCI} illiteracy.
\newblock \emph{Brain Topography}, 23\penalty0 (2):\penalty0 194--198., 2010.

\bibitem[Vishwanathan and Smola(2004)]{VIS04}
S.V.N. Vishwanathan and A.J. Smola.
\newblock {{Binet-Cauchy} kernels}.
\newblock In \emph{{Advances in Neural Information Processing Systems (NIPS)}},
  2004.

\bibitem[Wang et~al.(2006)Wang, Wang, and Feng]{WAN06}
L.~Wang, X.~Wang, and J.~Feng.
\newblock {Subspace distance analysis with application to adaptive Bayesian
  algorithm for face recognition}.
\newblock \emph{Pattern Recognition}, 39\penalty0 (3):\penalty0 456--464, 2006.

\bibitem[Welch(1974)]{WEL74}
L.~Welch.
\newblock Lower bounds on the maximum cross correlation of signals (corresp.).
\newblock \emph{IEEE Trans. Information Theory}, 20:\penalty0 397--399, 1974.

\bibitem[Wolf and Shashua(2003)]{WOL03}
L.~Wolf and A.~Shashua.
\newblock {Learning over sets using kernel principal angles}.
\newblock \emph{J. Mach. Learn. Res.}, 4:\penalty0 913--931, 2003.

\bibitem[Wong(1967)]{WON67}
Y.C. Wong.
\newblock {Differential geometry of Grassmann manifolds}.
\newblock \emph{PNAS}, 57\penalty0 (3):\penalty0 589--594, 1967.

\bibitem[Yaghoobi et~al.(2009)Yaghoobi, Daudet, and Davies]{Yaghoobi2009a}
M.~Yaghoobi, L.~Daudet, and M.E. Davies.
\newblock Parametric dictionary design for sparse coding.
\newblock \emph{IEEE Trans. Signal Processing}, 57:\penalty0 4800--4810, 2009.

\end{thebibliography}

\end{document}